%% file: iclr2026_conference.tex
\documentclass{article} % For LaTeX2e
\usepackage{iclr2026_conference,times}

\input{math_commands.tex}

\usepackage{hyperref}
\usepackage{url}
\usepackage{graphicx}
\usepackage{caption}
\usepackage{arydshln}
\usepackage{tcolorbox}
\usepackage[ruled,vlined]{algorithm2e}
\usepackage{algpseudocode}
\usepackage{multirow}
\usepackage{multirow}
\usepackage{arydshln}
\usepackage{booktabs}
\usepackage{arydshln}
\usepackage{wrapfig}
\usepackage{tabularx}
\usepackage{arydshln}
\usepackage{subcaption}
\usepackage{tikz}
\usepackage{hyperref}

\iclrpreprint

\definecolor{awcolor}{rgb}{1,0.4,0}

\newcommand{\model}{VideoJudge}

\title{VideoJudge: Bootstrapping Enables Scalable Supervision of MLLM-as-a-Judge for Video Understanding}

\author{
  Abdul Waheed\,
  ~~Zhen Wu\,
  ~~Dareen Alharthi\,
  ~~Seungone Kim\,
  ~~Bhiksha Raj \\
  Carnegie Mellon University \\
  \texttt{\normalsize \{abdulw,zhenwu,dalharth,seungonk,bhiksha\}@cs.cmu.edu}
}

\begin{document}

\maketitle

% \textcolor{red}{AW: Dummy abstract}

\begin{abstract}
Precisely evaluating video understanding models remains challenging: commonly used metrics such as BLEU, ROUGE, and BERTScore fail to capture the fineness of human judgment, while obtaining such judgments through manual evaluation is costly. Recent work has explored using large language models (LLMs) or multimodal LLMs (MLLMs) as evaluators, but their extension to video understanding remains relatively unexplored. In this work, we introduce VideoJudge, a 3B and 7B-sized MLLM judge specialized to evaluate outputs from video understanding models (\textit{i.e.}, text responses conditioned on videos). To train VideoJudge, our recipe builds on the interplay between a generator and an evaluator: the generator is prompted to produce responses conditioned on a target rating, and responses not matching the evaluator's rating are discarded. Across three out of four meta-evaluation benchmarks, VideoJudge-7B outperforms larger MLLM judge baselines such as Qwen2.5-VL (32B and 72B). Notably, we find that LLM judges (Qwen3) models perform worse than MLLM judges (Qwen2.5-VL) and long chain-of-thought reasoning does not improve performance, indicating that providing video inputs is crucial for evaluation of video understanding tasks.

\end{abstract}

% Include sections

% 1. Introduction
\input{sections/1_introduction}
\input{sections/2_related_work}
\input{sections/3_methodology}

% 4. Experiments
\input{sections/4_experiments}

% 5. Results
\input{sections/5_results}

% 6. Discussion
\input{sections/6_discussion}

% 7. Conclusion
\input{sections/7_conclusion}



% \clearpage
\bibliography{iclr2026_conference, custom}
\bibliographystyle{iclr2026_conference}

\appendix
% \section{Appendix}
\input{sections/8_appendix}

\end{document}

%% file: math_commands.tex
\usepackage{amsmath,amsfonts,bm}

\def\eqref#1{equation~\ref{#1}}
\def\1{\bm{1}}

\DeclareMathAlphabet{\mathsfit}{\encodingdefault}{\sfdefault}{m}{sl}
\SetMathAlphabet{\mathsfit}{bold}{\encodingdefault}{\sfdefault}{bx}{n}

%% file: sections/1_introduction.tex
\section{Introduction}\label{sec:introduction}

Recent advances in multimodal large language models (MLLMs) have significantly improved video captioning, question answering, and long-form video understanding across various domains. However, their progress poses a critical challenge: how to evaluate their outputs with reliability, interpretability, and at scale? Traditional reference-based metrics such as BLEU, ROUGE~\citep{lin-2004-rouge}, and BERTScore~\citep{zhang2020bertscoreevaluatingtextgeneration} struggle to capture semantic fidelity, contextual grounding, or task-specific reasoning. Moreover, in open-ended tasks where multiple valid answers exist, simple reference overlap can be misleading. Human evaluation is often considered the gold standard, but is expensive, slow to scale, and suffers from inter-annotator variability~\citep{liang2025evqascorefinegrainedmetricvideo}. 
A promising alternative is \textit{LLM-as-a-Judge}. By prompting or fine-tuning language models to assess responses, this paradigm has improved evaluation in text generation~\citep{zheng2023judgingllmasajudgemtbenchchatbot, kim2024prometheusinducingfinegrainedevaluation, gu2025surveyllmasajudge, li2024llmsasjudgescomprehensivesurveyllmbased} and more recently in vision–language tasks via \textit{MLLM-as-a-Judge}~\citep{chen2024mllmasajudgeassessingmultimodalllmasajudge, xiong2025llavacriticlearningevaluatemultimodal, lee2024prometheusvisionvisionlanguagemodeljudge}.

Applying MLLM-as-a-judge to video understanding remains underexplored, largely due to the temporal and multimodal complexity of video. Beyond this inherent difficulty, two broader limitations persist. First, the field lacks large-scale evaluation resources: there are no comprehensive datasets with human preference signals or standardized benchmarks for verifying alignment with human judgments. As a result, existing work either relies on proprietary models such as GPT-4 or GPT-4o~\citep{pu2025judgeanythingmllmjudge}, which lack transparency and reproducibility, or on small open-source MLLMs in zero-shot settings, which fall short of human-level reliability. Second, principled evaluation criteria are missing. Current (M)LLM-as-a-judge methods depend either on generic rubrics, which are often vague and brittle, or on manually authored rubrics, which cannot scale across tasks.

To address this gap, we introduce a framework to bootstrap data to train scalable video understanding evaluators. The framework has two key pillars. First, it automatically generates training data by producing candidate responses across a 1–5 rating scale, validating them with an evaluator model, and refining cases where predicted ratings diverge from expectations. These bootstrapped examples are then used to train both pointwise and pairwise judge models. Second, the same process enables the construction of new pointwise and pairwise meta-evaluation benchmarks, providing large-scale, high-quality resources for systematic comparison. In this way, our approach eliminates the need for costly human annotation while yielding both robust training data and standardized evaluation suites.

% Conventional (M)LLM-as-judge evaluation models rely either on generic rubrics or on manually designed instance-specific rubrics. Generic rubrics often yield unreliable judgments and are highly sensitive to small perturbations\cite{}, while scaling human-authored rubrics across instances is infeasible. To address this limitation, we train our judge model to automatically generate instance-specific rubrics at test time and then evaluate candidate responses against its own rubrics. Although this approach introduces additional computational overhead due to longer generations, it encourages the model to ground its decisions in explicit, context-specific criteria—making the evaluation more interpretable and, in turn, more reliable.
Second, we train MLLM judge models not only to predict ratings with explanations, but also to generate instance-specific rubrics at test time. This enables fine-grained evaluation that is both interpretable and anchored in explicit standards. Experiments in both pointwise and pairwise settings show that VideoJudge matches or surpasses much larger models while correlating more strongly with human ratings and demonstrating higher sample efficiency. 

In summary, our contributions are four-fold:
\begin{itemize}
    \item We introduce \textbf{VideoJudge}, the first bootstrapped framework for training scalable MLLM-based evaluators across diverse video understanding tasks.
    \item Train judge models that can not only assign ratings but also generate high-quality, instance-specific rubrics at inference time.
    \item We demonstrate that fine-tuned small models on the bootstrapped data can match or outperform much larger models in accuracy and alignment with human-specified ratings.
    \item We provide a suite of trained pointwise and pairwise judge models, meta-evaluation benchmarks, bootstrapped datasets, and other artifacts to support reproducible research in video understanding evaluation.
\end{itemize}

%% file: sections/2_related_work.tex
\section{Related Works}\label{sec:related_work}

\textbf{Video Understanding Models and Evaluation}
Recent advances in large language models have driven rapid extension into multimodal settings, where models jointly process and generate across text, image, audio, and video modalities~\citep{bai2025qwen2, chen2024expanding, wu2024next, xu2025qwen25omnitechnicalreport, zhao2025r1omniexplainableomnimultimodalemotion, chen2025janus, wu2025janus}. A growing line of work explores video understanding specifically, either by pretraining multimodal models with video–text corpora~\citep{zhang2024llavanextvideo, damonlpsg2023videollama, damonlpsg2024videollama2, damonlpsg2025videollama3, wang2025internvideo} or by instruction-tuning to align video representations with downstream tasks~\citep{zhang2024videoinstructiontuningsynthetic, zhang2024long}. These models are often evaluated using conventional, automatic metrics such as BLEU~\citep{papineni-etal-2002-bleu}, ROUGE~\citep{lin-2004-rouge}, and BERTScore~\citep{zhang2020bertscoreevaluatingtextgeneration}, which all assume the existence of reference answers. Human evaluation is also widely used but is costly and inconsistent. These limitations call for more principled automatic and semi-automatic approaches.

% are focused on extending them to support multimodal inputs~\cite{bai2025qwen2, chen2024expanding} and in many cases outputs~\cite{wu2024next, xu2025qwen25omnitechnicalreport, xu2025qwen25omnitechnicalreport, zhao2025r1omniexplainableomnimultimodalemotion, chen2025janus, wu2025janus}. These models often support multimodal understanding, including video understanding~\cite{}. Multimodal models that are natively trained for video understanding also can be continuously trained for video understanding tasks\cite{zhang2024llavanextvideo, zhang2024videoinstructiontuningsynthetic, damonlpsg2023videollama, damonlpsg2024videollama2, damonlpsg2025videollama3, wang2025internvideo, zhang2024long}. 

\textbf{LLM-as-Judge}
An alternative paradigm for evaluation leverages LLMs themselves as evaluators. Several works have investigated the viability of prompting powerful models such as GPT-4 to act as judges on text generation tasks~\citep{zheng2023judgingllmasajudgemtbenchchatbot, liu2023g, ye2023flask}. Beyond prompting, other efforts fine-tune open-weight models such as Llama-2~\citep{touvron2023llama} and Mistral~\citep{jiang2023mistral7b} to serve as reliable evaluators by distilling from GPT-4’s assessment trajectories~\citep{kim2023prometheus, kim2024prometheus, kim2025biggenbenchprincipledbenchmark}.  More recently, researchers have extended this line of work to multimodal settings. For example, \citet{chen2024mllm} examine whether multimodal LLMs can function as judges, while \citet{lee2024prometheus} explore fine-tuning open-weight models such as LLaVA-1.5~\citep{liu2024improved} to mimic the evaluation capability of proprietary MLLMs. Similarly, \citet{he2024videoscore} and~\citet{ku2024viescore} investigate the use of MLLMs as judges for text-to-image and text-to-video tasks. Together, these works highlight the promise of LLM-as-a-Judge for scalable evaluation, while underscoring the need to further test its robustness in video understanding.

%% file: sections/3_methodology.tex
\section{Methodology}\label{sec:methodology}

Our bootstrapping framework consists of a generator–evaluator pipeline that jointly synthesizes data and enforces quality control. The design draws inspiration from self-refinement approaches, where self-consistency~\citep{mitchell2022enhancingselfconsistencyperformancepretrained, wang2023selfconsistencyimproveschainthought, chen2023universalselfconsistencylargelanguage} and self-verification~\citep{weng2023largelanguagemodelsbetter} enhance LLM performance, and models adapt through verbal feedback~\citep{madaan2023selfrefineiterativerefinementselffeedback}. Our overall framework has two stages: (1) iterative bootstrapping to construct large-scale, fine-grained training data, and (2) fine-tuning judge models to generate ratings and instance-specific rubrics, which are evaluated under both pointwise and pairwise settings. Our framework is shown in Figure~\ref{fig:framework}.

\input{figures/framework}
% self-consistency~\citep{mitchell2022enhancingselfconsistencyperformancepretrained, wang2023selfconsistencyimproveschainthought, chen2023universalselfconsistencylargelanguage} and self-verification~\citep{weng2023largelanguagemodelsbetter} have been shown to improve the performance of large language models. Moreover, these models can, to some extent, adapt their responses based on verbal feedback~\citep{madaan2023selfrefineiterativerefinementselffeedback}. Drawing inspiration from the self-refinement framework, we use a generator–evaluator loop to iteratively construct high-quality training data for a video judge model. Our method has two components: (1) bootstrapping the training data and (2) training the judge model, followed by evaluation on a range of meta-evaluation benchmarks. Figure~\ref{fig:framework} provides an overview, and the following sections describe each component.

\subsection{Bootstrapping Process}
\label{subsec:bootstrapping}

We begin with seed data sourced from three large-scale video instruction–response datasets: VideoInstruct-100K~\citep{Maaz2023VideoChatGPT}, VCG-Plus-112K~\citep{Maaz2024VideoGPT+}, and VideoChat2-IT~\citep{2023videochat}. For VideoChat2-IT, which contains multi-turn dialogues, we retain only the first human–assistant exchange. The three corpora are merged and deduplicated at the instruction level for each video using a \textsc{MinHashLSH} index (128 permutations, Jaccard threshold $0.9$). From this deduplicated pool, we randomly sample $25\mathrm{K}$ examples, resulting in a corpus of triplets $(v, x, y^{*})$ where $v$ is a video, $x$ an instruction, and $y^{*}$ a gold-standard response.

% The instruction data in these corpora predominantly covers open-ended video understanding tasks, including question answering, captioning, instruction following, temporal and spatial reasoning, and multi-step inference grounded in video content. 

To transform this seed corpus into a training dataset for the evaluator, we iteratively generate and refine candidate responses for each $(v, x, y^{*})$ triplet. The process follows three stages: \emph{Initial Generation}, \emph{Feedback}, and \emph{Refinement}, described formally below.

\textbf{Initial Generation:}
For each instruction–video pair $(x, v)$ with gold response $y^{*}$, a generator model $G$ produces $N-1$ candidate responses, each intended to correspond to a rating $r \in \{1, \dots, N-1\}$ as shown in \ref{eq:initial_gen}. The gold response $y^{*}$ is included as the highest-rated response with rating $N$. 
\begin{equation}
y^{(r)}_{0} = G(p_{\text{gen}} \Vert v \Vert x \Vert y^{*}, r).
\label{eq:initial_gen}
\end{equation}

% \textbf{Feedback:}
% Each candidate $y^{(r)}_{t}$ is evaluated by an evaluator model $E$, which assigns a rating $\hat{r}$ and provides reasoning $f^{(r)}_{t}$ \ref{eq:feedback_f}. :

% \begin{equation}
% \hat{r}, f^{(r)}_{t} = E(p_{\text{eval}} \Vert v \Vert x \Vert y^{*} \Vert y^{(r)}_{t}).
% \label{eq:feedback_f}
% \end{equation}
% We compute the deviation between intended and assigned ratings as
% \begin{equation}
% \Delta^{(r)}_{t} = |r - \hat{r}|.
% \end{equation}
% Candidates with $\Delta^{(r)}_{t} \leq \alpha$ are accepted directly into the dataset.
\textbf{Feedback:}  
Each candidate response $y^{(r)}_{t}$ is evaluated by an evaluator model $E$, which assigns a rating $\hat{r}$ and provides reasoning $f^{(r)}_{t}$. We then compute the deviation between the intended rating $r$ and the evaluator’s rating $\hat{r}$ to determine whether the candidate should be accepted or refined. Candidates for which $\Delta^{(r)}_{t} \leq \alpha$ are accepted directly into the dataset. The evaluation process can be formalized as:

% \begin{align}
% \hat{r}, f^{(r)}_{t} &= E(p_{\text{eval}} \Vert v \Vert x \Vert y^{*} \Vert y^{(r)}_{t}) \label{eq:feedback_f} \\
% \Delta^{(r)}_{t} &= |r - \hat{r}| \label{eq:rating_deviation}
% \end{align}

\noindent
\begin{minipage}[t]{0.48\textwidth}
\begin{equation}
\hat{r}, f^{(r)}_{t} = E(p_{\text{eval}} \Vert v \Vert x \Vert y^{*} \Vert y^{(r)}_{t})
\label{eq:feedback_f}
\end{equation}
\end{minipage}
\hfill
\begin{minipage}[t]{0.48\textwidth}
\begin{equation}
\Delta^{(r)}_{t} = |r - \hat{r}|
\label{eq:rating_deviation}
\end{equation}
\end{minipage}

% \textbf{Refinement:}
% For candidates with $\Delta^{(r)}_{t} > \alpha$, we prompt the generator again using evaluator feedback:
% \begin{equation}
% y^{(r)}_{t+1} = G(p_{\text{ref}} \Vert v \Vert x \Vert y^{*} \Vert y^{(r)}_{t} \Vert f^{(r)}_{t}, r).
% \end{equation}
% This process repeats until convergence or a maximum of $T$ iterations.

\textbf{Refinement:}  
For candidates with a rating deviation $\Delta^{(r)}_{t} > \alpha$, the generator is prompted again using the evaluator’s feedback to improve the response. This iterative refinement continues until the candidate meets the acceptance criterion or a maximum of $T$ iterations is reached.The refinement step is formalized as:

\begin{equation}
y^{(r)}_{t+1} = G(p_{\text{ref}} \Vert v \Vert x \Vert y^{*} \Vert y^{(r)}_{t} \Vert f^{(r)}_{t}, r)
\label{eq:refinement}
\end{equation}

\textbf{Acceptance Criterion:}
A candidate response $y^{(r)}_{t}$ is added to the bootstrapped dataset if $|r - \hat{r}| \leq \alpha$. The final dataset, therefore, consists of $\{(v, x, y, r)\}$ triplets with aligned ratings. 

The complete process is outlined in Algorithm~\ref{alg:bootstrap}. Using this pipeline, we bootstrap pointwise data with $N=5$, where each instruction is paired with five responses rated from 5 to 1. Representative examples are shown in Table~\ref{tab:bootstrapped_examples} in Appendix~\ref{appsubsec:boostrapping}.

\subsection{Model Training}\label{subsec:training}

We use the bootstrapped dataset to train pointwise and pairwise evaluator models. The dataset is structured as $\mathcal{D} = \{(v_i, x_i, y_i, t_i)\}_{i=1}^{M}$, where $v_i$ denotes the video, $x_i$ the instruction, $y_i$ a candidate response (or a response pair in the pairwise setting), and $t_i$ the associated target annotation, such as a rating or a preference label. The evaluator model $E_\theta$ is trained end-to-end to autoregressively generate the target sequence $t_i$ conditioned on $(v_i, x_i, y_i)$, with the standard negative log-likelihood over tokens serving as the objective: 
$$
\mathcal{L}(\theta) = - \frac{1}{M} \sum_{i=1}^M \sum_{j=1}^{|t_i|} \log P_\theta\big(t_{i,j} \mid t_{i,<j}, v_i, x_i, y_i \big),
$$
where $t_{i,j}$ denotes the $j$-th token of $t_i$. This loss is applied to both pointwise and pairwise models. In the pointwise setting, the model produces intermediate reasoning within \texttt{<thinking></thinking>} followed by a scalar rating in \texttt{<score></score>}, and it can optionally generate task-specific rubrics in \texttt{<rubric></rubric>} before reasoning and evaluation. In the pairwise setting, the model outputs its decision within \texttt{<answer></answer>} based on a pair of candidate responses.

% \aw{Maybe we need to add parameters details of generators and evaluators here only.}

% \input{sections/algorithm}

%% file: figures/framework.tex
\begin{figure*}[hbtp]
    \centering
    \includegraphics[width=\linewidth]{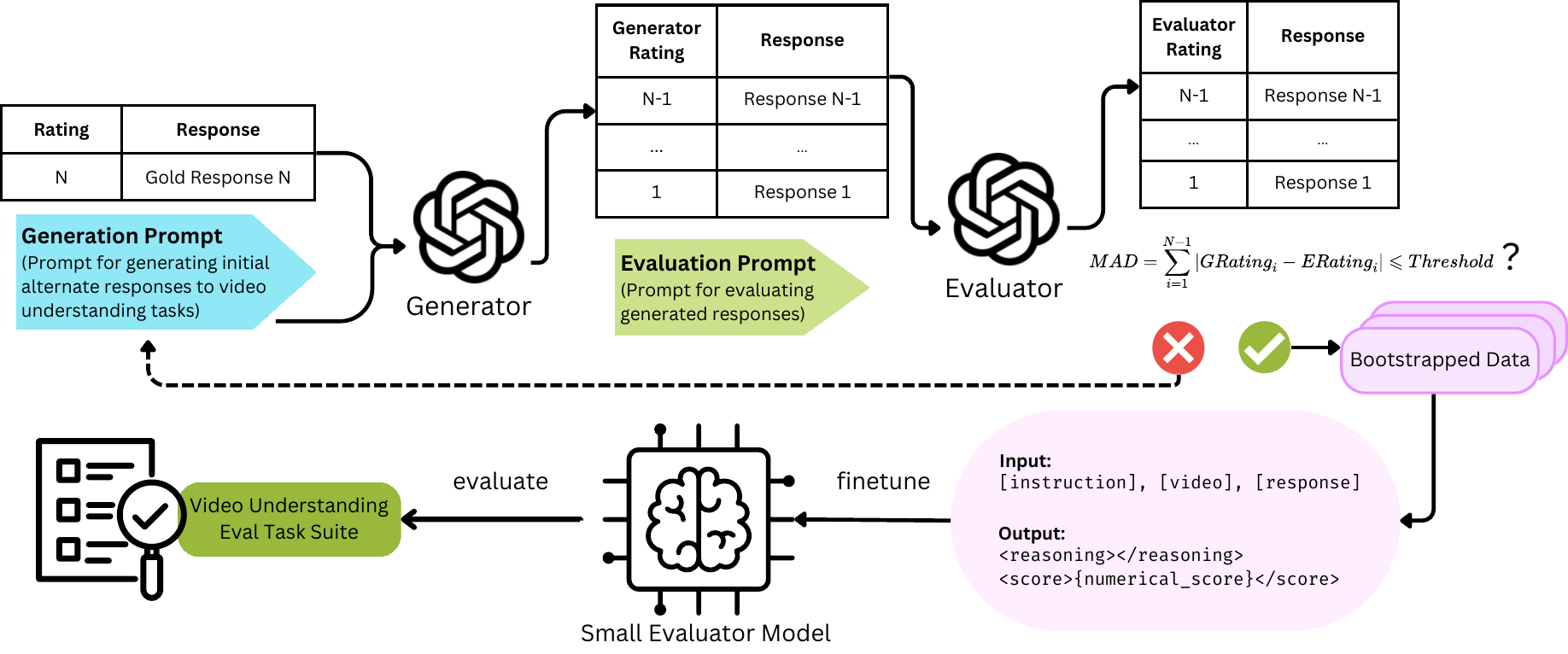} 
    \caption{Overview of our bootstrapping framework for training scalable video evaluators. 
    A \emph{generator} first produces candidate responses for a 1 to $N\!-\!1$ rating scale ($N=5$) for each video–instruction pair. 
    These responses are then scored by an \emph{evaluator}, and only candidates whose ratings align with expectations are retained. 
    Through an iterative refinement loop, mismatched responses are revised until they satisfy the acceptance criterion. 
    The resulting bootstrapped dataset provides high-quality supervision signals, which we use to fine-tune compact VideoJudge models.}
    \label{fig:framework}
\end{figure*}

%% file: sections/4_experiments.tex
\section{Experiments}\label{sec:experiments}

We bootstrap pointwise data starting from $25\mathrm{K}$ seed video instructions-response pairs. After the bootstrapping process, we retain only instructions with at least five responses (one for each rating), yielding $103{,}825$ examples across $20{,}765$ unique video–instruction pairs. 
% This expands the initial seed into a large-scale corpus of over $100{,}000$ examples covering diverse videos, instructions, model outputs, and evaluation labels.  

We construct pairwise supervision by forming response pairs where the higher-rated output is chosen as preferred. Due to computational limitations and while keeping the setting identical, we randomly sample $50\%$ of all possible pairs, resulting in $103{,}825$ pairwise training examples. Both pointwise and pairwise judge models are trained on these bootstrapped datasets. In the \textit{pointwise} setting, the model takes a video, instruction, and candidate response as input, and is trained to produce a reasoning trace followed by a quality score. We further train a judge model to first generate an instruction-specific evaluation rubric, which is then applied when scoring, ensuring that evaluations are grounded in context-specific criteria. In the \textit{pairwise} setting, each instruction is paired with two candidate responses\footnote{To avoid positional bias, the order of responses is randomized during both training and evaluation.}, and the model is trained to identify the preferred response.

% The ground-truth label corresponds to the response with the higher rating. To mitigate order bias, we randomize the order of responses in training data.  

% the input consists of a video, instruction, and response. The model is trained to generate a reasoning trace followed by a score (with rubrics in some experiments).  
% For \textit{pairwise training}, each instruction is paired with two candidate responses, and the model is trained to identify which response is better. The ground-truth label corresponds to the response with the higher rating. To mitigate order bias, we randomize the order of responses in training data.  

% We provide further details on the experimental setup in the following subsections.
  
\subsection{Baselines}\label{subsec:baselines}
% We evaluate a broad range of models, including both unimodal language models and multimodal video-language models, to benchmark their ability as judges. Unimodal models are tested using detailed video descriptions as proxies for visual input, while video models directly process the video content. This setup allows us to compare how different model classes handle the judging task.

We compare VideoJudge against both unimodal language models and multimodal video–language models. Unimodal baselines are given detailed video descriptions generated by Qwen2.5-VL-72B as proxies for visual input, while video models process video frames directly.

\textbf{Unimodal Models}  
% Since unimodal language models cannot directly process visual inputs, we generate detailed video descriptions using large and powerful models (e.g., Qwen2.5-VL-72B) and use them as a proxy for the original video. 
Each unimodal model is provided with the description, instruction, and candidate response, and is prompted to generate a reasoning trace followed by a score. We consider Qwen3~\citep{yang2025qwen3technicalreport} family of models from 0.6B to 14B, and also enable the ``thinking mode'' of smaller models (up to 4B) to test whether extended reasoning sequences enhance judging ability~\citep{chan2025j1exploringsimpletesttime, kim2025scalingevaluationtimecomputereasoning, zhou2025evaluatingjudgesevaluatorsjetts}. 

% Prior work has shown that increasing test-time compute by generating longer sequences can improve performance across various tasks~\cite{}, a strategy often referred to as \emph{thinking mode}. To explore this effect, we also enable the `thinking mode` of Qwen models for smaller configurations (up to 4B) to test whether extended reasoning sequences enhance judging ability.

\textbf {Video Models}  
We evaluate Qwen2.5-VL(3B–72B)~\citep{bai2025qwen25vltechnicalreport} along with other recent video–language models, including LLaVA-Next (7B)~\citep{zhang2024llavanextvideo}, VideoR1 (7B)~\citep{feng2025videor1reinforcingvideoreasoning}, and LLaVA-OneVision~\citep{li2024llavaonevisioneasyvisualtask}. In our preliminary experiments we find that several models—such as VideoLLaMA3-7B~\citep{zhang2025videollama3frontiermultimodal}, VideoChat-Flash~\citep{li2024videochatflash}, Keye-VL~\citep{yang2025kwaikeyevl15technical}, and SmolVLM2~\citep{marafioti2025smolvlm}—frequently failed to follow instructions or produce valid scores under the same evaluation setup. Consequently, we exclude them from our main results.

\subsection{Evaluation}\label{subsec:evaluation}

% In a pointwise setup, we measure whether a model can assign meaningful scores to individual responses, while pairwise evaluation tests whether it can consistently prefer the better of two responses to the same instruction. Together, these approaches provide a comprehensive view of model performance across diverse benchmarks.

\paragraph{Pointwise} Each video--instruction--response triplet is evaluated independently, with the model producing a reasoning trace followed by a rating on a 1--5 scale. We construct two meta-evaluation benchmarks, VideoJudgeLLaVA-MetaEval and VideoJudgeVCG-MetaEval, by sourcing seed instruction data from LLaVA-Video~\citep{zhang2024videoinstructiontuningsynthetic} and VideoChatGPT~\footnote{\url{https://huggingface.co/datasets/lmms-lab/VideoChatGPT}}, then generating additional responses via our bootstrapping pipeline (Algorithm~\ref{alg:bootstrap}) with threshold $0$. We report correlation and error metrics, as well as divergence error. We further evaluate on Vatex-Eval~\citep{shi2022emscoreevaluatingvideocaptioning}, which contains multiple human judgments aggregated into continuous ground-truth scores, emphasizing ranking- and separation-based measures to capture preference consistency. We also use LongVideoBench~\citep{wu2024longvideobenchbenchmarklongcontextinterleaved} for long-form multiple-choice evaluation by rating correct versus distractor answers, reporting both the average score gap (\emph{Delta}) and \emph{pairwise superiority}.

\paragraph{Pairwise}  The judge models compare two candidate responses for the same video--instruction and select the preferred one. We use VideoAutoArena~\citep{luo2025videoautoarenaautomatedarenaevaluating}, where human preferences serve as ground truth. From our pointwise evaluation data, we also construct VideoJudge-Pairwise by pairing responses with different ratings and treating the higher-rated response as correct, measuring accuracy against this derived ground truth. To probe more subtle distinctions, we create VideoJudge-Pairwise-H focusing on challenging 2-vs.-3 cases: we sample 250 such pairs, collect annotations from two human evaluators, and retain only those with full agreement, yielding over 200 pairs with human preference ground truth. We report the accuracy score for all pairwise evaluations.

\paragraph{Experimental Setup}  
% \aw{Need to add all the details and create a hyperparameter table.}
All models are trained and evaluated under identical hyperparameter settings to ensure fairness. We use \textit{full finetuning} in \textit{BF16} precision with a maximum sequence length of 128K tokens with fps rate of 1 with max number of frames 60 for training and 180 during evaluation. We train all models for 2 epochs with a batch size of 16. The learning rate is set to $2\!\times\!10^{-7}$ with cosine decay, a warmup ratio of 0.03, weight decay of 0, and gradient clipping at 1. We provide key hyperparameters and other implementation details in Appendix~\ref{appsubsec:hyperparameters}

% We provide our experimental setup in Appendix~

%% file: sections/5_results.tex
\section{Data Evaluation}\label{sec:data_generation}
We evaluate the bootstrapped data to ensure the quality and that it provides meaningful and reliable supervision. Our data evaluation has two parts: automatic checks to assess the relative quality of the responses, and human evaluation to validate correctness and preference alignment. Together, these evaluations confirm that the generated data is of sufficient quality for training and benchmarking.

\subsection{Automatic Evaluation}
During our bootstrapping process, we prompt the generator model to produce candidate responses for different ratings by progressively degrading the quality according to the specified score. A natural proxy to verify that the generated dataset adheres to this design is to assess whether response quality indeed declines as we move from higher to lower ratings. To this end, we compute BERTScore and BLEU using the gold response as reference and the generated candidates (ratings 4--1) as hypotheses. The results, presented in Figure~\ref{fig:bert_bleu_combined}, exhibit a clear monotonic degradation: BERTScore decreases from $91.1$ (5--4) to $86.9$ (5--1), while BLEU drops from $11.0$ to $3.0$. This consistent downward trend confirms that the generator reliably produces responses of progressively lower quality, validating the effectiveness of our controlled response generation process.
\input{figures/automatic_data_eval}

\subsection{Human Evaluation}

% \textbf{Pairwise:} 
 We construct our pairwise data by sampling response pairs with different ratings for the same instruction, choosing the higher-rated one as preferred. In practice, we found that generator–evaluator disagreements were most frequent around ratings 2 and 3, even after incorporating the feedback loop. To focus on these harder cases, we restricted human evaluation to pairs with ratings 2 vs.\ 3.  We sample 250 such examples, each containing the video, a detailed description, the instruction, and two candidate responses, and asked two annotators to select the preferred response. Results are shown in Table~\ref{tab:humaneval_results}. Agreement between annotators is high (94.8\% with Cohen’s $\kappa$ of 89.5), and both annotators achieved over 92\% correctness relative to the gold preference. The preference distribution shows a mild bias toward response \texttt{b}, but error analysis indicates only 4.4\% cases where both annotators agreed on the wrong response and 5.2\% where they disagreed on a correct one. Overall, the study confirms that the generated pairwise data is consistent and reliable, even in the most ambiguous rating regions. We provide detailed metrics of our human evaluation study in Table~\ref{tab:humaneval_results} and examples in Table~\ref{tab:annotator_comparison}, Appendix~\ref{appsubsec:human_evaluation}.

\section{Results and Discussion}\label{sec:results}
We train Qwen2.5-VL (3B, 7B) models for both pointwise and pairwise evaluation under identical settings. We evaluate various baselines and our trained judge models across a suite of meta-evaluation benchmarks. We report pointwise evaluation results in Table~\ref{tab:results_pointwise} and pairwise results in Table~\ref{tab:results_pairwise}. Our findings show that bootstrapped supervision enables smaller models to reach, and in some cases surpass, the judgment reliability and accuracy of much larger ($\sim$10$\times$) general-purpose models. We discuss our findings in subsequent sections.

\subsection{Pointwise Evaluation}
We evaluate all models in the pointwise setup, where each system is required 
to produce a scalar score. We use the identical prompt decoding parameter across the different models. Unimodal models provide a useful reference point. Across Qwen3 variants, 
we find that performance on VideoJudgeLLaVA and VideoJudgeVCG is reasonably 
strong, but VATEX remains challenging with consistently high error and poor 
calibration. LongVideoBench is more demanding: while unimodal models achieve 
non-trivial PSup values, the margin between correct and distractor responses 
is narrow, reflecting the difficulty of capturing temporal dependencies from 
text-only signals. Thinking mode further improves the 0.6B model, showing that 
explicit reasoning steps help even in the pointwise regime. However, enabling unimodal models to perform judgment requires high-quality, detailed descriptions, often generated by powerful models such as Qwen2.5-VL-72B or GPT-4o-mini. Thus, the cost of description generation should be included in the overall cost.
% However, it's important to note that to enable unimodal models to perform judgment, we need high-quality, detailed descriptions often generated by powerful models such as Qwen2.5-VL-72B or GPT-4o-mini. Hence, the cost of the description generation process should be accounted for in the overall cost. 

Video-language model baselines such as LLaVA-NeXT, OneVision, and Video-R1 perform competitively on VideoJudgeLLaVA and VideoJudgeVCG, achieving correlations in the 0.66--0.77 range with relatively low error values, on par with or better than several unimodal Qwen3 baselines. However, their performance degrades substantially on LongVideoBench, where both PSup and $\Delta$(C--D) drop sharply (e.g., LLaVA-NeXT: 0.59 / 0.45, Video-R1: 0.60 / 0.54), underscoring the difficulty of long-context temporal reasoning. In contrast, the Qwen2.5-VL series scales more robustly: larger variants (32B, 72B) show consistent improvements across all four benchmarks, achieving higher correlations and stronger $\Delta$(C--D) values on LongVideoBench.

Our trained VideoJudge models deliver consistently strong performance across all evaluation settings, establishing a new standard for video-language judgment. On VideoJudgeLLaVA and VideoJudgeVCG, both VideoJudge-3B and VideoJudge-7B achieve correlations that not only match but in several cases surpass those of substantially larger baselines such as Qwen2.5-VL-32B/72B, while also outperforming post-trained baseline video-language systems including LLaVA-NeXT, OneVision, and Video-R1. Beyond short-context benchmarks, VATEX results further underscore the advantages of feedback-guided training, with our models exhibiting lower error and improved calibration, yielding predictions that are both accurate and well-grounded. The most pronounced gains are observed on LongVideoBench, a challenging benchmark for temporal reasoning, where existing models degrade sharply but VideoJudge models maintain high PSup and $\Delta$(C--D) scores. These improvements demonstrate that feedback-based supervision imparts a capacity for consistent and temporally coherent evaluation, enabling VideoJudge models to deliver stable judgments even in extended and complex video contexts. Overall, these results show that rubric-supervised judges match or surpass larger video-language models, providing a scalable and principled approach to reliable multimodal evaluation.

\input{tables/results_pointwise}

\input{tables/results_pointwisewithrubric}

\paragraph{Training Judge Models to Generate Instance-Specific Rubrics at Test Time} In our setup, we first synthesize training rubrics and then train the model to (i) generate a rubric for each instance, (ii) reason with the rubric, and (iii) output an integer score. This approach enables scalable, rubric-driven evaluation tailored to individual examples. For computational feasibility, we train \textit{Qwen2.5-VL-3B} on 10\% of total pointwise data and evaluate on 1{,}000 examples sampled from VideoJudgeLLaVA and VideoJudgeVCG. We report the results in Table~\ref{tab:results_pointwisewithrubric}. The rubric generation prompt is shown in Figure~\ref{fig:rubric_generation}, and the training/evaluation prompt is provided in Figure~\ref{fig:train_eval_pointwisewithrubric} in Appendix~\ref{appsubsec:prompts}.

Our results show that \textit{VideoJudgeR-3B}, trained to generate instance-specific rubrics, substantially improves over the 3B and 7B baselines. It reduces error (MAE 0.59 vs.~1.15, RMSE 1.05 vs.~1.56) and achieves correlations above 73, comparable to the much larger 32B and 72B base models. This demonstrates that rubric-driven supervision can close most of the performance gap without scaling model size, yielding evaluations that are both more reliable and more interpretable.

\paragraph{Evaluating the Quality of Generated Rubrics}  

% While rubric-driven supervision improves model performance, it is equally important to verify whether the generated rubrics themselves are meaningful and useful for guiding evaluation. High-quality rubrics should articulate explicit, context-specific criteria, whereas poor rubrics risk being vague or generic. To assess rubric quality, we use two complementary methods: \textbf{LLM-as-Judge}, where \textit{GPT-4o-mini} ($temperature=0$) selects the better rubric between two candidates, and human evaluation, where 300 rubric pairs per model are judged by three annotators, with outcomes aggregated by unanimous and majority vote. This dual setup enables us to measure alignment with both automated LLM judgments and human preferences. 
While rubric-driven supervision improves model performance, it is also important to verify whether the rubrics themselves are meaningful and useful for evaluation. High-quality rubrics should specify explicit, context-specific criteria, whereas poor ones risk being vague or generic. To assess rubric quality, we use two methods: \textbf{LLM-as-Judge}, where \textit{GPT-4o-mini} ($temperature=0$) selects the better rubric between two candidates, and human evaluation, where 300 rubric pairs per model are judged by three annotators, with outcomes aggregated by unanimous (as shown in Figure~\ref{fig:videojudge_winrate}) or majority vote (Figure~\ref{fig:videojudge_winrate_additional}). This dual setup measures alignment with both automated LLM judgments and human preferences.

\input{figures/videojudge_winrate}

Our results show that \textit{VideoJudgeR-3B} produces substantially higher-quality rubrics than the 3B, 7B, and 32B baselines across all settings, with large margins under both unanimous and majority human judgments and even stronger gains in LLM-as-Judge evaluation. Against stronger models, it continues to hold an edge: in the LLM-as-Judge setup, VideoJudgeR-3B achieves a 92.7\% win rate against GPT-4o-mini and 71.3\% against Qwen-72B, consistently maintaining above 50\% win rate across all evaluation settings. These findings demonstrate that instance-specific rubric generation enables a compact 3B model to outperform much larger models while producing rubrics preferred by both humans and strong LLM judges. We provide example rubrics generated by different models in Table~\ref{tab:rubrics_example} in Appendix~\ref{appsubsec:training_and_evaluation}.

\input{tables/results_pairwise}

\subsection{Pairwise Evaluation}
We next train and evaluate models in the pairwise setting, where the task is to prefer the better of two responses to the same video--instruction pair. This setup directly captures relative quality and aligns closely with human preference judgments. As before, we assess both base models and our trained VideoJudge models, with and without feedback, across VideoAutoArena (VAA), VideoJudge (VJ), and VideoJudge-Human (VJ-H). Results are summarized in Table~\ref{tab:results_pairwise}. VideoJudge models consistently outperform their backbone baselines across all benchmarks. Notably, \textbf{VideoJudge-3B} achieves 94.0 on VJ and 89.45 on VJ-H (w/ feedback), far surpassing Qwen2.5-VL-3B (82.6 / 85.23) and even outperforming much larger models such as Qwen2.5-VL-32B and 72B in several cases. \textbf{VideoJudge-7B} further improves performance, attaining 98.6 on VJ and 93.67 on VJ-H. These results highlight that our bootstrapped enables smaller models to match or exceed the reliability of much larger video-language systems. Feedback provides consistent gains for 3B and 7B baselines, while its benefits diminish for larger models, and in VideoJudge models, it yields mixed but benchmark-specific effects.

\paragraph{How Many Frames Are Enough for an Effective Video Judge}  
We study the effect of \texttt{maxframes} on video judgment performance, as it controls the temporal context available for evaluation. Too few frames risk omitting critical evidence, while excessively large values increase computation without proportional benefit. To analyze this tradeoff, we vary \texttt{maxframes} during training (30–500, evaluation fixed at 180) and separately during evaluation (30–180, training fixed at 60). This design isolates the role of temporal coverage in both training and inference.  

When varied during training, VideoJudge shows consistent gains from larger \texttt{maxframes}. Correlations with ground truth rating increase steadily up to $\sim$240 frames (exceeding 0.7), while RMSE and MAE decline. Beyond this point, improvements plateau, suggesting diminishing returns despite higher cost. In evaluation, increasing \texttt{maxframes} at inference improves correlation and reduces error up to $\sim$120 frames, after which performance saturates.  

Overall, these results indicate that moderate to large temporal context is crucial for effective judgment. Training benefits from covering up to 240 frames, while at evaluation, modest values (around 120) suffice to capture most relevant evidence. Thus, carefully chosen \texttt{maxframes} can balance accuracy and efficiency, strengthening temporal grounding without unnecessary cost.

\input{figures/temperature_spearman}
\paragraph{Decoding Temperature} 
We study the effect of decoding temperature on sampling reliability, as it directly controls the trade-off between determinism and diversity. Its impact matters for evaluation models, where unstable sampling can cause inconsistent judgments.

Figure~\ref{fig:temperature_spearman} (other metrics in Figure~\ref{fig:temperature_others} in Appendix~\ref{appsubsec:results}) reports the pointwise performance of the base Qwen2.5-VL-3B and its VideoJudge-trained counterpart across a range of temperatures. The base model degrades steadily as temperature increases, with Spearman correlation falling from 0.56 at $T=0.0$ to 0.42 at $T=1.0$, accompanied by higher error rates and more invalid outputs. In contrast, the VideoJudge model remains robust and even benefits from higher temperatures, peaking at a correlation of 0.73 and achieving the lowest MAE of 0.69. These findings indicate that while naïve sampling destabilizes alignment, rubric-guided training both stabilizes performance and allows models to exploit higher-temperature decoding to better capture distributional richness. Such robustness is especially valuable in practice, where non-deterministic decoding is often preferred to promote diversity.

% \paragraph{Instruction Following}

%% file: figures/automatic_data_eval.tex
% \begin{wrapfigure}[16]{r}{0.55\textwidth}
%     \centering
%     \vspace{-30pt}
%     \includegraphics[width=0.4\textwidth]{assets/bert_bleu_combined.pdf}
%     \vspace{-10pt}
%     \caption{Comparison of \textbf{BERTScore} (left axis, blue) and \textbf{BLEU} (right axis, orange) across rating pairs. 
%     BERTScore remains consistently high while BLEU drops sharply with larger rating gaps.}
%     \label{fig:bert_bleu_combined}
%     \vspace{-5pt}
% \end{wrapfigure}

\begin{wrapfigure}[19]{r}{0.45\textwidth}
    \centering
    % \vspace{-20pt}
    \includegraphics[width=0.45\textwidth]{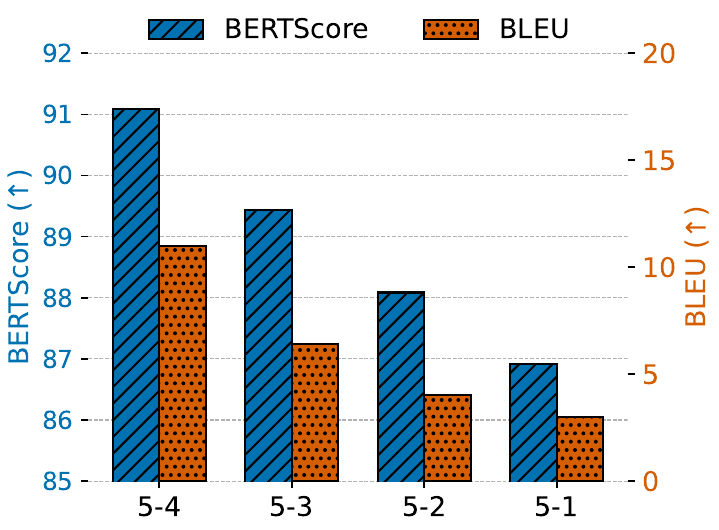}
    % \vspace{-8pt}
    \caption{The monotonic decrease in BERTScore and BLEU score demonstrates that our proposed framework is capable of producing responses with controlled quality.}
    \label{fig:bert_bleu_combined}
    \vspace{-12pt}
\end{wrapfigure}

%% file: tables/results_pointwise.tex
\begin{table*}[!htbp]
\caption{Benchmark results across VideoJudgeLLaVa, VideoJudgeVCG, VATEX, and LongVidB. Metrics: RMSE/MAE (error), S/P (Spearman/Pearson correlation), ECE (calibration), PSup/\,$\Delta$(C-D) (preference).}
\centering
\scriptsize
\setlength{\tabcolsep}{1.5pt}
\renewcommand{\arraystretch}{1.1}
\begin{tabular}{@{}l@{\hspace{8pt}}lcccccccccccc@{}}
\toprule
\textbf{Model} & & \multicolumn{4}{c}{\textbf{VideoJudgeLLaVa}} & \multicolumn{4}{c}{\textbf{VideoJudgeVCG}} & \multicolumn{2}{c}{\textbf{VATEX}} & \multicolumn{2}{c}{\textbf{LongVidB}} \\
& & RMSE$_{\downarrow}$ & MAE$_{\downarrow}$ & S$_{\uparrow}$ & P$_{\uparrow}$ & RMSE$_{\downarrow}$ & MAE$_{\downarrow}$ & S$_{\uparrow}$ & P$_{\uparrow}$ & RMSE$_{\downarrow}$ & ECE$_{\downarrow}$ & PSup$_{\uparrow}$ & $\Delta$(C-D)$_{\uparrow}$ \\
\midrule
\multirow{5}{*}{\rotatebox[origin=c]{90}{\textit{Unimodal}}} 
& Qwen3-0.6B  & 1.20 & 0.92 & 0.64 & 0.63 & 1.20 & 0.92 & 0.66 & 0.65 & 1.44 & 0.84 & 0.58 & 0.17 \\
& Qwen3-1.7B  & 0.97 & 0.60 & 0.78 & 0.78 & 1.33 & 0.85 & 0.61 & 0.60 & 2.18 & 0.86 & 0.64 & 0.71 \\
& Qwen3-4B    & 0.96 & 0.61 & 0.72 & 0.74 & 1.22 & 0.79 & 0.58 & 0.59 & 2.42 & 0.97 & 0.67 & 0.69 \\
& Qwen3-8B    & 0.97 & 0.61 & 0.73 & 0.74 & 1.19 & 0.76 & 0.61 & 0.61 & 2.06 & 0.94 & 0.64 & 0.53 \\
& Qwen3-14B   & 1.09 & 0.65 & 0.69 & 0.69 & 1.29 & 0.83 & 0.58 & 0.58 & 2.38 & 0.96 & 0.63 & 0.57 \\
\midrule
\multirow{3}{*}{\rotatebox[origin=c]{90}{\shortstack{\textit{Unimodal} \\ \textit{(thinking)}}}}
\\
& Qwen3-0.6B  & 1.12 & 0.81 & 0.64 & 0.64 & 1.31 & 0.98 & 0.59 & 0.59 & 1.98 & 0.89 & 0.62 & 0.63 \\
& Qwen3-1.7B  & 1.00 & 0.70 & 0.73 & 0.75 & 1.34 & 0.94 & 0.56 & 0.58 & 1.99 & 0.82 & 0.65 & 0.86 \\
& Qwen3-4B    & 1.02 & 0.69 & 0.68 & 0.70 & 1.33 & 0.90 & 0.51 & 0.52 & 2.12 & 0.92 & 0.65 & 0.69 \\
\midrule

\multirow{7}{*}{\rotatebox[origin=c]{90}{\textit{Video Models}}}
& LLaVA-NeXT-7B    & 1.06 & 0.72 & 0.67 & 0.66 & 1.06 & \textbf{0.67} & 0.70 & 0.69 & 1.65 & 0.84 & 0.59 & 0.45 \\
& LLaVA-OneVision  & 1.01 & 0.71 & 0.77 & 0.75 & 1.00 & 0.69 & \textbf{0.77} & 0.76 & 1.52 & 0.78 & 0.64 & 0.83 \\
& Video-R1-7B      & 1.07 & 0.67 & 0.73 & 0.73 & 1.74 & 1.21 & 0.46 & 0.47 & 1.87 & 0.72 & 0.6 & 0.54 \\
& Qwen2.5-VL-3B    & 1.31 & 0.94 & 0.63 & 0.63 & 1.58 & 1.12 & 0.51 & 0.52 & 2.27 & 0.85 & 0.56 & 0.20 \\
& Qwen2.5-VL-7B    & 0.92 & 0.61 & 0.77 & 0.76 & 1.22 & 0.76 & 0.65 & 0.65 & 2.36 & 0.88 & 0.57 & 0.35 \\
& Qwen2.5-VL-32B   & \textbf{0.87} & 0.59 & 0.80 & 0.79 & 1.05 & 0.75 & 0.69 & 0.70 & 1.43 & 0.81 & \textbf{0.73} & 1.08 \\
& Qwen2.5-VL-72B   & \textbf{0.87} & 0.61 & 0.80 & 0.81 & \textbf{0.98} & 0.69 & 0.76 & \textbf{0.77} & 1.40 & 0.79 & 0.71 & 1.06 \\
\midrule
\multirow{2}{*}{\rotatebox[origin=c]{90}{\textit{Ours}}}
& VideoJudge-3B    & 1.07 & 0.61 & \textbf{0.82} & \textbf{0.82} & 1.59 & 1.06 & 0.59 & 0.63 & \textbf{1.33} & \textbf{0.63} & 0.61 & 0.70 \\
& VideoJudge-7B    & 0.96 & \textbf{0.52} & 0.78 & 0.80 & 1.20 & 0.72 & 0.74 & 0.76 & 1.46 & 0.64 & 0.66 & \textbf{1.16} \\
\bottomrule
\end{tabular}
\label{tab:results_pointwise}
\end{table*}

%% file: tables/results_pointwisewithrubric.tex
\begin{wraptable}[14]{r}{0.45\textwidth} % reduced width + height
\centering
\setlength{\tabcolsep}{2.5pt} % tighter columns
\renewcommand{\arraystretch}{1.0} % shrink row height
\caption{\scriptsize Divergence errors and correlation metrics 
(P = Pearson, S = Spearman) for zero-shot base models and 
the VideoJudgeR-3B model trained to generate instance-specific 
rubrics at test time. All models are prompted to produce rubrics 
together with reasoning and a score.}
\label{tab:results_pointwisewithrubric}
\scriptsize % shrink font for table
\begin{tabular}{lcccc}
\toprule
\textbf{Model} & \textbf{MAE$_{\downarrow}$} & \textbf{RMSE$_{\downarrow}$} & \textbf{P$_{\uparrow}$} & \textbf{S$_{\uparrow}$} \\
\midrule
Qwen2.5-VL-3B   & 1.15 & 1.56 & 37.85 & 37.96 \\
Qwen2.5-VL-7B   & 0.86 & 1.22 & 57.09 & 57.26 \\
Qwen2.5-VL-32B  & 0.59 & 0.86 & 78.59 & 80.21 \\
Qwen2.5-VL-72B  & 0.54 & 0.87 & 78.10 & 78.61 \\
\midrule
VideoJudgeR-3B  & 0.59 & 1.05 & 73.96 & 74.16 \\
\bottomrule
\end{tabular}
\end{wraptable}

% \begin{figure}[t]
%     \centering
%     \begin{minipage}{0.48\textwidth}
%         \centering
%         \setlength{\tabcolsep}{3.5pt}
%         \renewcommand{\arraystretch}{0.95}
%         \begin{tabular}{lcccc}
%         \toprule
%         \textbf{Model} & \textbf{MAE$_{\downarrow}$} & \textbf{RMSE$_{\downarrow}$} & \textbf{P$_{\uparrow}$} & \textbf{S$_{\uparrow}$} \\
%         \midrule
%         Qwen2.5-VL-3B   & 1.15 & 1.56 & 37.85 & 37.96 \\
%         Qwen2.5-VL-7B   & 0.86 & 1.22 & 57.09 & 57.26 \\
%         Qwen2.5-VL-32B  & 0.59 & 0.86 & 78.59 & 80.21 \\
%         Qwen2.5-VL-72B  & 0.54 & 0.87 & 78.10 & 78.61 \\
%         \midrule
%         VideoJudgeR-3B  & 0.59 & 1.05 & 73.96 & 74.16 \\
%         \bottomrule
%         \end{tabular}
%         \captionsetup{width=\textwidth}
%         \captionof{table}{Divergence errors and correlation metrics (P = Pearson, S = Spearman) for zero-shot base models and the VideoJudgeR-3B model trained to generate instance-specific rubrics at test time. All models are prompted to produce rubrics together with reasoning and a score.}
%         \label{tab:results_pointwisewithrubric}
%     \end{minipage}
%     \hfill
%     \begin{minipage}{0.48\textwidth}
%         \centering
%         \includegraphics[width=\linewidth]{assets/videojudge_winrate_human_unanimous.pdf}
%         \caption{Win rates of \textit{VideoJudge-3B} against other models in pairwise rubric preference evaluation.}
%         \label{fig:videojudge_winrate}
%     \end{minipage}
% \end{figure}

%% file: figures/videojudge_winrate.tex
\begin{wrapfigure}[16]{r}{0.52\textwidth}
    \centering
    \vspace{-1.0em}
    \includegraphics[width=\linewidth]{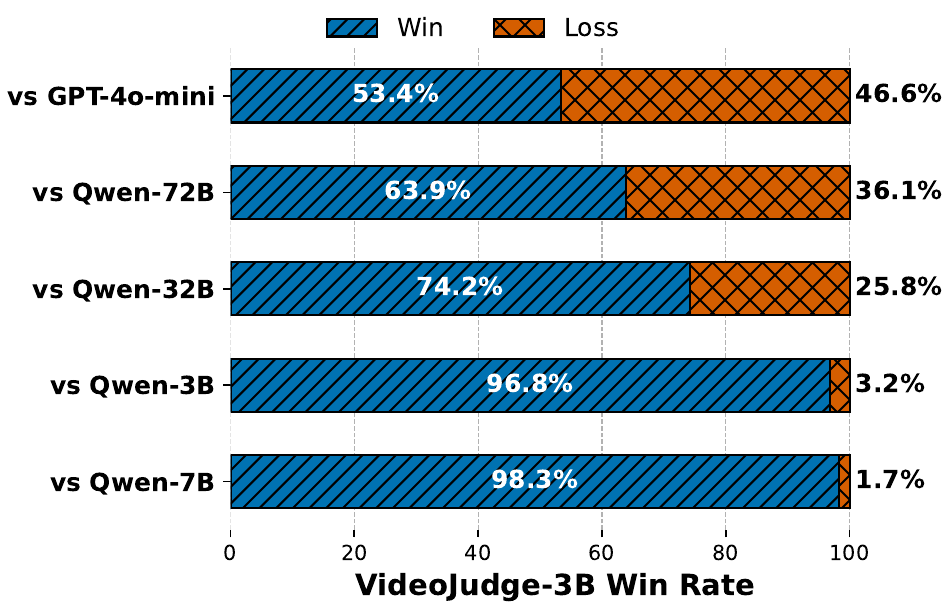}
    \vspace{-1.2em}
    \caption{Win rates from human evaluations comparing \textit{VideoJudge-3B} against other models.}
    \label{fig:videojudge_winrate}
\end{wrapfigure}

%% file: tables/results_pairwise.tex
% \begin{wraptable}[10]{r}{0.62\textwidth}
% \centering
% \setlength{\tabcolsep}{3pt}
% \renewcommand{\arraystretch}{1.2}
% \resizebox{\linewidth}{!}{%
% \begin{tabular}{ccccccccc}
% \toprule
% \multirow{2}{*}{Model} 
%   & \multicolumn{2}{c}{VAA} 
%   & \multicolumn{2}{c}{VJ} 
%   & \multicolumn{2}{c}{VJ-H} \\
% \cmidrule(lr){2-3}\cmidrule(lr){4-5}\cmidrule(lr){6-7}
%   & w/ FB & w/o FB & w/ FB & w/o FB & w/ FB & w/o FB \\
% \midrule
% Qwen2.5-VL-3B   & 54.90 & 52.16 & 82.60 & 75.00 & 85.23 & 81.01 \\
% Qwen2.5-VL-7B   & 75.29 & 71.37 & 89.00 & 84.60 & 89.03 & 82.28 \\
% Qwen2.5-VL-32B  & 80.78 & \textbf{90.59} & 91.20 & 91.20 & 92.83 & 90.72 \\
% Qwen2.5-VL-72B  & \textbf{89.80} & \underline{89.80} & 94.00 & 93.20 & \textbf{94.51} & \textbf{93.25} \\
% \midrule
% VideoJudge-3B   & 71.76 & 64.71 & \underline{94.00} & \underline{95.80} & 89.45 & 90.72 \\
% VideoJudge-7B   & \underline{85.49} & 87.45 & \textbf{95.60} & \textbf{98.60} & \underline{93.67} & \textbf{93.25} \\
% \bottomrule
% \end{tabular}}
% \caption{Accuracy scores ($\uparrow$) of zero-shot base models and \model~on pairwise meta-evaluation benchmarks. 
% Abbreviations: VAA = VideoAutoArena, VJ = VideoJudge, VJ-H = VideoJudge-Human, 
% w/ FB = with feedback, w/o FB = without feedback.}
% \label{tab:results_pairwise}
% \end{wraptable}

% \vspace{-2.5em} % Adjust this value as needed
\begin{wraptable}[18]{r}{0.62\textwidth}
\centering
\setlength{\tabcolsep}{3pt}
\renewcommand{\arraystretch}{1.2}
\caption{Accuracy scores ($\uparrow$) of zero-shot base models and \model~on pairwise meta-evaluation benchmarks. 
Abbreviations: VAA = VideoAutoArena, VJ = VideoJudge, VJ-H = VideoJudge-Human, 
w/ FB = with feedback, w/o FB = without feedback.}
\label{tab:results_pairwise}
\resizebox{\linewidth}{!}{%
\begin{tabular}{ccccccccc}
\toprule
\multirow{2}{*}{Model} 
  & \multicolumn{2}{c}{VAA} 
  & \multicolumn{2}{c}{VJ} 
  & \multicolumn{2}{c}{VJ-H} \\
\cmidrule(lr){2-3}\cmidrule(lr){4-5}\cmidrule(lr){6-7}
  & w/ FB & w/o FB & w/ FB & w/o FB & w/ FB & w/o FB \\
\midrule
Qwen2.5-VL-3B   & 54.90 & 52.16 & 82.60 & 75.00 & 85.23 & 81.01 \\
Qwen2.5-VL-7B   & 75.29 & 71.37 & 89.00 & 84.60 & 89.03 & 82.28 \\
Qwen2.5-VL-32B  & 80.78 & \textbf{90.59} & 91.20 & 91.20 & 92.83 & 90.72 \\
Qwen2.5-VL-72B  & \textbf{89.80} & \underline{89.80} & 94.00 & 93.20 & \textbf{94.51} & \textbf{93.25} \\
\midrule
VideoJudge-3B   & 71.76 & 64.71 & \underline{94.00} & \underline{95.80} & 89.45 & 90.72 \\
VideoJudge-7B   & \underline{85.49} & 87.45 & \textbf{95.60} & \textbf{98.60} & \underline{93.67} & \textbf{93.25} \\
\bottomrule
\end{tabular}}
% \caption{Accuracy scores ($\uparrow$) of zero-shot base models and \model~on pairwise meta-evaluation benchmarks. 
% Abbreviations: VAA = VideoAutoArena, VJ = VideoJudge, VJ-H = VideoJudge-Human, 
% w/ FB = with feedback, w/o FB = without feedback.}
% \label{tab:results_pairwise}
\end{wraptable}

%% file: figures/temperature_spearman.tex
\begin{wrapfigure}[20]{r}{0.45\textwidth}
    \centering
    % \vspace{-8pt}
    \includegraphics[width=\linewidth]{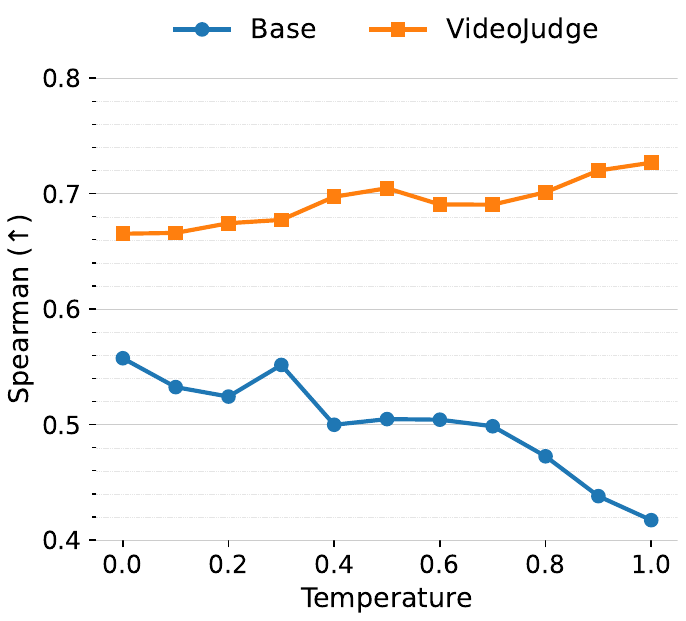}
    \vspace{-8pt}
    \caption{Spearman correlation across temperatures for base and our video models.}
    \label{fig:temperature_spearman}
\end{wrapfigure}

%% file: sections/6_discussion.tex
% \section{Discussion and Analysis}\label{sec:discussion}

%% file: sections/7_conclusion.tex
\section{Conclusion}\label{sec:conclusion}

% We introduce VideoJudge, a bootstrapping framework for training MLLM-based evaluators for video understanding tasks. Our iterative generator-evaluator pipeline creates over 100,000 training examples without human annotation and enables models to generate both ratings and instance-specific rubrics at test time . Experiments show that VideoJudge-3B matches models up to 10× larger, while VideoJudge-7B consistently outperforms larger baselines across benchmarks. The models correlate strongly with human judgments and maintain performance in long-context scenarios where baselines degrade. VideoJudgeR-3B produces rubrics preferred by human annotators and LLM judges. We release curated benchmarks, datasets, and models to support reproducible multimodal evaluation research.

We introduce VideoJudge, a bootstrapping framework for training MLLM-based evaluators specialized for video understanding. Our approach addresses the lack of evaluation resources with human preference signals and principled evaluation criteria for video understanding. The core contribution lies in an iterative generator-evaluator pipeline that synthesizes training data and enforces quality control, creating over 100,000 training examples without costly human annotation. We fine-tune judge models to generate both ratings and instance-specific rubrics at test time, enabling interpretable evaluations anchored in explicit criteria grounded in the specific instruction and video content. Our experiments demonstrate that fine-tuned 3B and 7B VideoJudge models match or outperform much larger baselines in accuracy and alignment with human ratings. VideoJudge-3B achieves comparable performance to models up to 10× larger, while VideoJudge-7B consistently outperforms larger video-language models across multiple benchmarks. VideoJudgeR-3B produces rubrics preferred by both human annotators and LLM judges while maintaining performance comparable to much larger base models. By releasing curated meta-evaluation benchmarks, bootstrapped datasets, and trained models, we provide essential resources for reproducible multimodal evaluation research. The bootstrapping methodology is general and could extend to other modalities beyond video understanding.

% \section{Reproducibility Statement}\label{sec:reproducibility}
% We are committed to ensuring the reproducibility of our results. To this end, we have anonymously released the model checkpoints and datasets used for training and evaluation on \href{https://huggingface.co/xyzasdfghjkl123456}{HuggingFace}. We an index of released artifacts in Table~\ref{tab:artifacts} in Appendix~\ref{appsubsec:reproducibility}. All remaining code, data, checkpoints, and other related artifacts will be made available either during the review process or upon acceptance.

%% file: sections/8_appendix.tex
\section{Appendix}\label{appendix}

\subsection{Related Work}\label{appsubsec:related_work}
\paragraph{Evaluation Benchmarks}
% Video understanding models are evaluated on a wide range of static benchmarks, ranging from reasoning~\cite{} to downstream tasks~\cite{}. Earlier work includes evaluating the models for tasks like video captioning and correlating the performance with human evaluators~\cite{wang2020vatexlargescalehighqualitymultilingual, shi2022emscoreevaluatingvideocaptioning} 

% While many meta-evaluation benchmarks exist both for unimodal settings and visual understanding~\cite{xiong2025llavacriticlearningevaluatemultimodal} 

Video understanding models are evaluated on a wide range of benchmarks spanning different tasks~\citep{sanders2024survey}. For captioning, datasets such as MSR-VTT~\citep{xu2016msr}, VATEX~\citep{wang2020vatexlargescalehighqualitymultilingual}, and HowTo100M~\citep{miech2019howto100m} provide large-scale paired video–text data, with evaluation often relying on reference-based metrics or correlation with human judgments~\citep{shi2022emscoreevaluatingvideocaptioning}. For action recognition, datasets like ActivityNet offer a large-scale benchmark covering hundreds of activity categories with temporal annotations~\citep{caba2015activitynet}. For video question answering, datasets such as TVQA~\citep{lei2018tvqa} and NEXT-QA~\citep{xiao2021next} require models to integrate visual content with natural language reasoning over semantically complex or temporally extended video segments. In parallel, meta-evaluation benchmarks have been proposed to measure the reliability of evaluators themselves, both in unimodal and multimodal settings. Examples include RewardBench~\citep{lambert2024rewardbench} and MM-EVAL~\citep{son2024mm} in the unimodal domain, and multimodal resources such as VATEX EVAL~\citep{shi2022emscoreevaluatingvideocaptioning}, VLRewardBench~\citep{li2024vlrewardbench}, Judge Anything~\citep{pu2025judge}, and LLaVA-Critic~\citep{xiong2025llavacriticlearningevaluatemultimodal}. 

\subsection{Dataset}
Here we provide more details about the videos used in our study. More specifically, we provide duration statistics of the videos along with the nature instruction data in Table~\ref{tab:video_durations_and_description}. 

\input{tables/dataset_stats}

\subsection{Prompts}\label{appsubsec:prompts}
In this section, we provide a comprehensive list of prompts that we use in our study.

\subsubsection{Bootstrapping}

\paragraph{Response Generation Prompt}\label{generation_prompt}  
The prompt used to generate candidate responses is provided in Figure~\ref{fig:candidate_response_generation}.  
\input{figures/prompts/gen}

\paragraph{Response Evaluation Prompt}\label{evaluation_prompt}  
The prompt for evaluating candidate responses during the bootstrapping stage is shown in Figure~\ref{fig:candidate_response_generation}.  
\input{figures/prompts/eval}

\paragraph{Response Regeneration from Feedback}\label{feedback_prompt}  
After the initial round of response generation and evaluation, we measure the difference between the generator’s self-assigned rating and the evaluator’s rating. This numerical gap is then incorporated into feedback, which guides response regeneration. The corresponding prompt is shown in Figure~\ref{fig:regeneration_with_feedback}.  
\input{figures/prompts/feedback}

\subsubsection{Training and Evaluation}\label{appsubsec:training_and_evaluation}
\paragraph{Pointwise} For pointwise evaluation, we prompt the model to produce a reasoning sequence followed by a scalar score, as illustrated in Figure~\ref{fig:train_eval_pointwise}.

\input{figures/prompts/train_eval_pointwise}

\paragraph{Rubric Generation}
\input{figures/prompts/rubric_gen}
We employ \emph{GPT-4o-mini} to construct evaluation rubrics conditioned on the instruction, the video description (serving as a proxy for the video content), and the gold standard response.

We provide example rubrics generated by different models in Table~\ref{tab:rubrics_example}

\input{tables/rubric_examples}

\paragraph{Pointwise Training, Evaluation, and Rubric Generation}
\input{figures/prompts/train_eval_pointwise}
\input{figures/prompts/train_eval_pointwisewithrubric}
\input{figures/prompts/rubric_eval}
We use the prompt in Figure~\ref{fig:train_eval_pointwise} to train VideoJudge models and to evaluate responses in a pointwise setup. The prompt in Figure~\ref{fig:train_eval_pointwisewithrubric} extends this by enabling models to both train and evaluate in a pointwise setting while also generating rubrics at test time. Finally, Figure~\ref{fig:rubric_comparison} shows the prompt used to evaluate rubrics produced by different models.

\paragraph{Pairwise training and evaluation.} 
\input{figures/prompts/train_eval_pairwise}
\input{figures/prompts/train_eval_pairwisewithfeedback}
We provide the prompt to train and evaluate models in a pairwise setup without feedback in Figure~\ref{fig:train_eval_pairwise} and with feedback generation in Figure~\ref{fig:train_eval_pairwisewithfeedback}.

\subsection{Boostrapping Process}\label{appsubsec:boostrapping}
\input{sections/algorithm}

\subsection{Evaluation Data}\label{appsubsec:evaluation}
\input{tables/bootstrapped_data_example}

\subsection{Hyperparameters}\label{appsubsec:hyperparameters}
In Table~\ref{tab:hyperparams} we provide a detailed list of hyperparameters we use in our experiments. We use Qwen2.5-VL~\footnote{\url{https://github.com/QwenLM/Qwen2.5-VL}} as training framework vLLM~\citep{kwon2023efficientmemorymanagementlarge} for evaluation. We keep all other parameters as default until stated otherwise. 
\input{tables/hyperparameters}

\subsection{Human Evaluation}\label{appsubsec:human_evaluation}
\subsubsection{Pairwise}
\input{tables/pairwise_humaneval}

\input{tables/pairwise_humaneval_examples}

\subsection{Results}\label{appsubsec:results}

\paragraph{Rubric Generation}
\input{figures/videojudge_winrate_additional}

\paragraph{Human Evaluation of Generated Rubrics}

\begin{figure}[htb]

\centering
\includegraphics[width=0.45\textwidth]{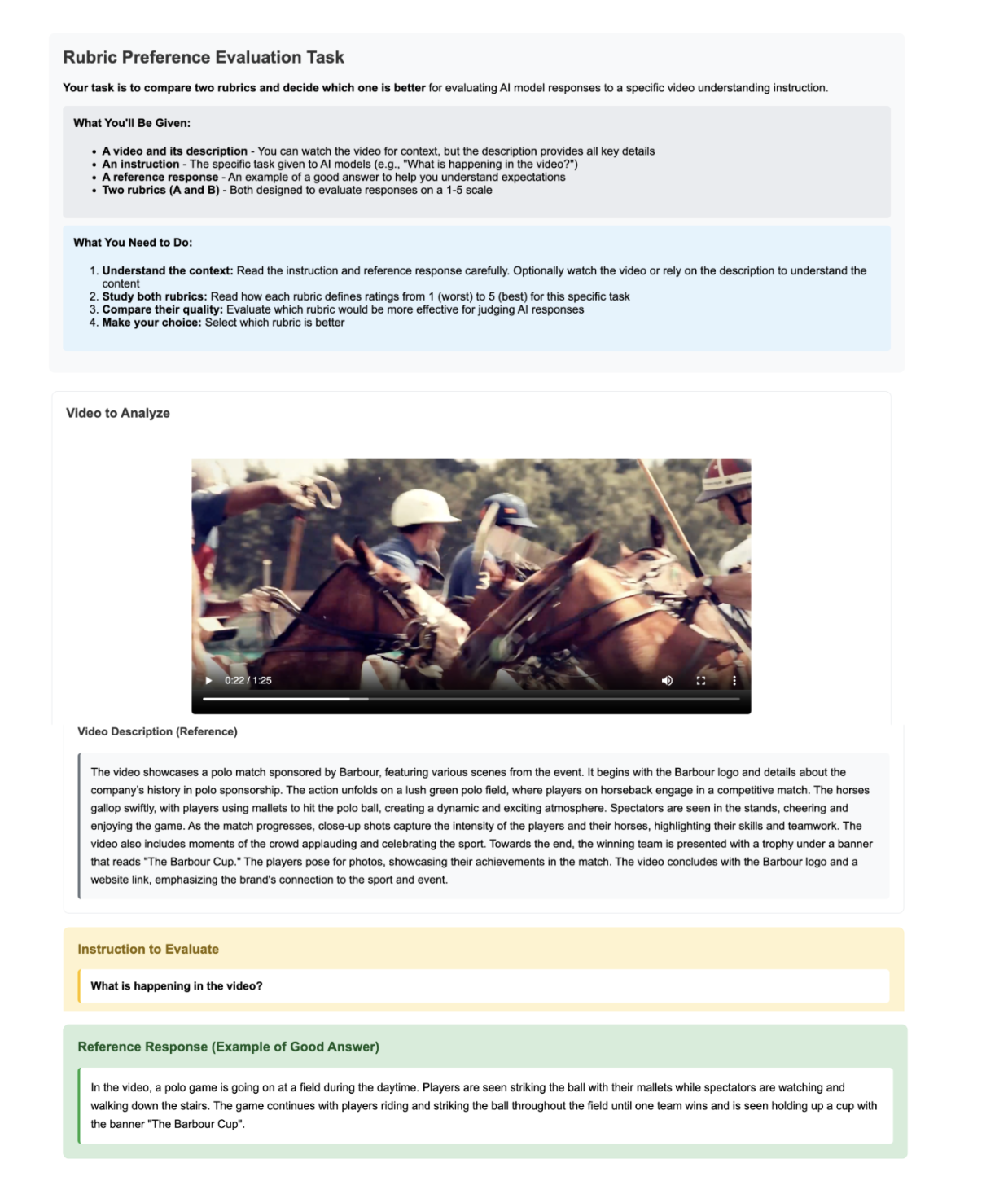}%
\hspace{-0.5cm}
\includegraphics[width=0.48\textwidth]{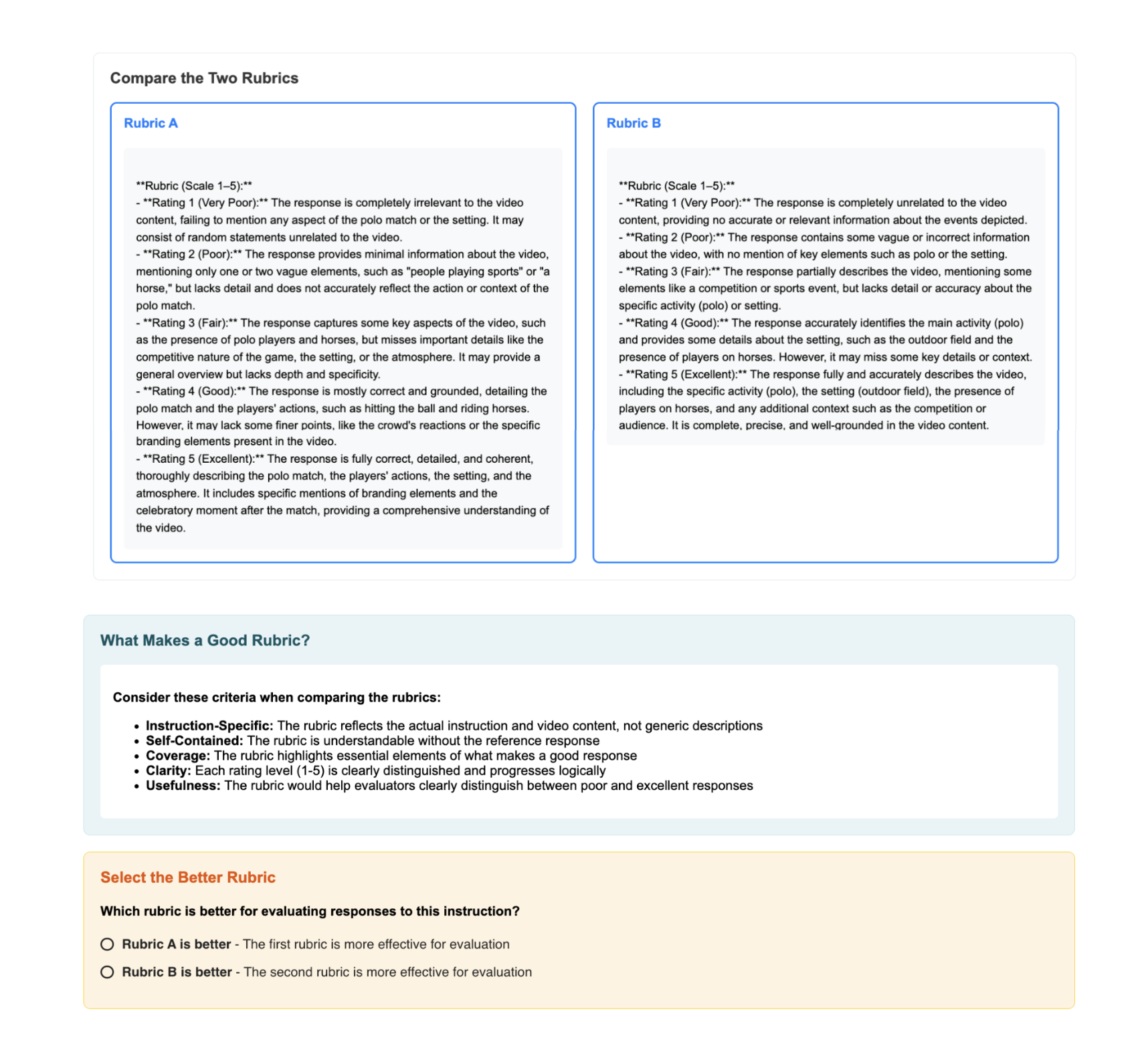}
\caption{Example of the Human Evaluation MTurk Interface. Annotators were provided with a video, its description, the instruction, a reference response, and two candidate rubrics. They compared the rubrics and selected the one they considered more effective for evaluating AI-generated responses.}
\label{fig:performance_combined}
\end{figure}

We conduct a human evaluation study on Amazon Mechanical Turk to compare the quality of rubrics generated for video–instruction–response triplets. The evaluation dataset consists of 300 randomly sampled rubrics for 6 models, including VideoJudge-3B (which we train). For each example, we generated rubrics using VideoJudge as one option and compared them against rubrics produced by five other models: GPT-4-mini, Qwen-3B, Qwen-7B, Qwen-32B, and Qwen-72B. Annotators were given the video along with its description, the instructions to the models, a reference response illustrating a good answer, and two candidate rubrics (A and B), each defined on a 1–5 scale. Their task was to compare the two rubrics and select the one they considered more effective for evaluating AI-generated responses to the given instruction. Each rubric pair was assessed independently by three annotators. The full annotation framework, including the instructions provided and a representative Human Intelligence Task (HIT) example, is shown in Figure~\ref{fig:performance_combined}.

% \begin{figure}[htb]
% \centering
% \includegraphics[width=0.8\textwidth]{figures/human_eval/response_human_eval.png}
% \caption{Human evaluation response analysis}
% \label{fig:response_eval}
% \end{figure}

\paragraph{Decoding Temperature} We provide other metrics for decoding at different temperature in Table~\ref{fig:temperature_others}.
\input{figures/temperature_others}

\paragraph{Number of Frames} We provide other metrics for max frames ablation in Table~\ref{fig:maxframes_others}.
\input{figures/maxframes_others}

%% file: tables/dataset_stats.tex
\begin{table}[ht!]
\centering
\setlength{\tabcolsep}{4pt} % tighter spacing
\renewcommand{\arraystretch}{1.05}
\caption{Video duration statistics (in seconds) across evaluation datasets, sorted by number of unique videos. Count indicates unique videos considered after deduplication.}
\label{tab:video_durations_and_description}
\resizebox{\linewidth}{!}{%
\begin{tabular}{lcccccl}
\toprule
\textbf{Duration/Dataset} & \textbf{Count} & \textbf{Min} & \textbf{Max} & \textbf{Mean} & \textbf{Median} & \textbf{Remark} \\
\midrule
VideoJudge-RS-20K         & 9469  & 2.1  & 745.4  & 117.6 & 114.4 & 20K-scale VideoJudge preference dataset \\
VideoJudgeLLaVA-MetaEval  & 3038  & 5.0  & 341.8  & 18.3  & 9.9   & VideoJudge benchmark with LLaVA prompts \\
VatexEval-MetaEval        & 2340  & 1.9  & 7180.0 & 167.5 & 81.4  & VATEX evaluation split (multilingual captions) \\
VideoJudgeVCG-MetaEval    & 499   & 3.0  & 238.0  & 108.0 & 98.7  & VideoJudge benchmark with VCG prompts \\
LongVideoBench-MetaEval   & 280   & 8.0  & 297.1  & 56.0  & 16.8  & Long-context video reasoning benchmark \\
VideoAutoArena-Preference & 241   & 8.0  & 3291.4 & 433.6 & 60.7  & Preference pairs from VideoAutoArena \\
\bottomrule
\end{tabular}}
\end{table}

%% file: figures/prompts/gen.tex
\begin{figure*}[h]
    \centering
    \tcbset{colframe=black, colback=gray!10, arc=5mm, title={Candidate Response Generation Prompt}}
    \begin{tcolorbox}
    \scriptsize
    You are provided with a \textbf{detailed video description}, a gold standard \textbf{response} rated 5 (perfectly accurate, highest quality), and a corresponding \textbf{instruction} for a \textbf{video understanding task}. This task may include \textbf{video captioning}, \textbf{video question answering}, \textbf{video instruction following}, \textbf{temporal action localization}, or any other open-ended \textbf{video understanding} scenario. \\
    
    Your task is to generate four additional \textbf{responses} that simulate progressively lower-quality outputs for the same \textbf{instruction}. Each generated \textbf{response} should correspond to a quality rating from \textbf{4} to \textbf{1}, where \textbf{Rating 5} is the provided gold standard and \textbf{Ratings 4 through 1} represent decreasing quality. \\
    
    As the rating decreases, the \textbf{responses} should reflect increasing levels of degradation, including \textbf{hallucinations}, \textbf{omissions}, \textbf{irrelevant information}, \textbf{logical inconsistencies}, or \textbf{grammatical issues}. All generated \textbf{responses} must remain similar in length to the gold standard and maintain the expected task format (e.g., caption, answer, instruction). Do \textbf{not} simply truncate the gold response --- simulate \textbf{realistic and meaningful degradation} in quality across levels. Use the \textbf{video description} to ground the correctness of the response content. \\
    
    \textbf{Task:} \\
    Generate four degraded \textbf{responses} corresponding to quality ratings \textbf{4} through \textbf{1}, based on the gold standard \textbf{response} (Rating 5). Follow the rating guidelines strictly. \\
    
    \textbf{Rating Guidelines:} \\
    - \textbf{Rating 4}: Mostly accurate with minor issues. Preserves the key meaning with small lapses in detail or precision. \\
    - \textbf{Rating 3}: Partially correct. Conveys the general idea but includes noticeable errors, omissions, or misinterpretations. \\
    - \textbf{Rating 2}: Poor alignment. Contains serious flaws and only loosely relates to the instruction or video. \\
    - \textbf{Rating 1}: Unrelated or incorrect. Fails to reflect the video or follow the instruction meaningfully. \\
    
    \textbf{Example:} \\
    \textbf{Instruction:} ``Can you give a brief summary of the video content?'' \\
    \textbf{Video Description:} ``A large group of people are participating in an aerobics class inside a spacious indoor venue...'' \\
    \textbf{Gold Standard Response (Rating 5):} ``The video is about a group of people doing a step exercise class, with some confusion among the participants. The focus is on the main woman, who stops at the end and walks towards the camera.'' \\
    \textbf{Rating 4:} ``The video shows a group of people in a step exercise class, with some participants appearing confused during the workout. The main woman stops exercising at the end and moves toward the camera.'' \\
    \textbf{Rating 3:} ``The video features people doing an aerobics class with stepping movements and some coordination issues. There's a woman who seems to be leading the group, and she approaches the camera at some point during the session.'' \\
    \textbf{Rating 2:} ``The video shows several people in what appears to be a dance or fitness routine taking place indoors. A woman in the group walks around and comes closer to where the video is being filmed, while others continue moving in the background.'' \\
    \textbf{Rating 1:} ``The video depicts a cooking demonstration where a chef wearing workout clothes explains different healthy recipes to a small audience in what looks like a kitchen or dining area, with people standing around tables.'' \\ 
    
    \textbf{Output Format:} \\
    Your final output \textbf{must} be a valid JSON object with exactly the following keys: \texttt{"rating\_4"}, \texttt{"rating\_3"}, \texttt{"rating\_2"}, and \texttt{"rating\_1"}. Each value should be the generated \textbf{response} corresponding to that quality rating. Do \textbf{not} include any additional commentary, formatting, or explanation outside the JSON object. \\
    
    \textbf{Input:} \\
    \textbf{Instruction:} \begin{verbatim}{instruction}\end{verbatim}
    \textbf{Video Description:} \begin{verbatim}{video_description}\end{verbatim}
    \textbf{Gold Standard Response (Rating 5):} \begin{verbatim}{gold_standard_response}\end{verbatim}
    
    \vspace{2mm}
    \textbf{Output (according to the JSON schema):} \\
    \begin{verbatim}
    {
      "rating_4": "[Generated response]",
      "rating_3": "[Generated response]",
      "rating_2": "[Generated response]",
      "rating_1": "[Generated response]"
    }
    \end{verbatim}
    \end{tcolorbox}
    \caption{LLM-as-judge prompt for generating degraded responses at different quality levels based on a gold standard.}
    \label{fig:candidate_response_generation}
\end{figure*}

%% file: figures/prompts/eval.tex
\begin{figure*}[h]
    \centering
    \tcbset{colframe=black, colback=gray!10, arc=5mm, title={Candidate Response Evaluation Prompt}}
    \begin{tcolorbox}
    \scriptsize
    You are provided with a detailed \textbf{video description}, an \textbf{instruction}, a gold standard \textbf{response} rated 5 (perfectly accurate, highest quality), and a set of candidate \textbf{responses} for a \textbf{video understanding task}. This task may include \textbf{video captioning}, \textbf{question answering}, \textbf{instruction following}, \textbf{temporal action localization}, or any other open-ended \textbf{video understanding} scenario. \\
    
    \textbf{Task:} \\
    Your task is to \textbf{evaluate each candidate response} and assign an \textbf{evaluation rating} from \textbf{1} to \textbf{4}. While the gold standard \textbf{response} can serve as a reference for what an ideal \textbf{response} looks like, your evaluation should primarily focus on how well each candidate \textbf{response} fulfills the task defined by the \textbf{instruction}, given the provided \textbf{video description}. Focus your reasoning on identifying what is \textbf{incorrect}, \textbf{missing}, or \textbf{misleading} in the \textbf{response} itself. \\
    
    \textbf{Evaluation Steps:} 
    \begin{enumerate}
        \item Read the \textbf{instruction} carefully to understand the intended task. 
        \item Refer to the gold standard \textbf{response} (rated 5) as a reference for correctness and completeness. 
        \item Use the provided \textbf{video description} as the factual basis for evaluating all responses. 
        \item For each candidate \textbf{response}: identify \textbf{inaccuracies}, \textbf{omissions}, \textbf{hallucinations}, or \textbf{irrelevant content}, and evaluate alignment with the \textbf{instruction} and \textbf{video description}. 
        \item Assign an integer score from \textbf{1--4} to each candidate \textbf{response}. 
        \item Provide reasoning for each rating, focusing on specific aspects that affect alignment with the \textbf{instruction} and \textbf{video description}.
    \end{enumerate}
    
    \textbf{Rating Guidelines for Evaluation:} \\
    - \textbf{Rating 4}: Mostly accurate with minor issues. Preserves key meaning with small lapses. \\
    - \textbf{Rating 3}: Partially correct. General idea conveyed but with noticeable errors/omissions. \\
    - \textbf{Rating 2}: Poor alignment. Serious flaws, loosely related to task or video. \\
    - \textbf{Rating 1}: Unrelated or incorrect. Fails to reflect the video or follow the instruction. \\
    
    \textbf{Example:} \\
    \textbf{Instruction:} ``Can you give a brief summary of the video content?'' \\
    \textbf{Video Description:} ``A large group of people are participating in an aerobics class inside a spacious indoor venue...'' \\
    \textbf{Gold Standard Response (Rating 5):} ``The video is about a group of people doing a step exercise class, with some confusion among the participants. The focus is on the main woman, who stops at the end and walks towards the camera.'' \\
    \textbf{Rating 4:} ``The video shows a group of people in a step exercise class, with some participants appearing confused during the workout. The main woman stops exercising at the end and moves toward the camera.'' \\
    \textbf{Rating 3:} ``The video features people doing an aerobics class with stepping movements and some coordination issues. There's a woman who seems to be leading the group, and she approaches the camera at some point during the session.'' \\
    \textbf{Rating 2:} ``The video shows several people in what appears to be a dance or fitness routine taking place indoors. A woman in the group walks around and comes closer to where the video is being filmed, while others continue moving in the background.'' \\
    \textbf{Rating 1:} ``The video depicts a cooking demonstration where a chef wearing workout clothes explains different healthy recipes to a small audience in what looks like a kitchen or dining area, with people standing around tables.'' \\
    
    \textbf{Output Format:} \\
    Your final output \textbf{must} follow the exact JSON schema provided. Do \textbf{not} include any additional formatting, comments, or text outside of the JSON object. \\
    
    \textbf{Input:} \\
    \textbf{Instruction:} \begin{verbatim}{instruction}\end{verbatim}
    \textbf{Video Description:} \begin{verbatim}{video_description}\end{verbatim}
    \textbf{Gold Standard Response (Rating 5):} \begin{verbatim}{gold_standard_response}\end{verbatim}
    \textbf{Generated Responses:} \begin{verbatim}{generated_responses}\end{verbatim}
    
    \vspace{2mm}
    \textbf{Output (according to the JSON schema):} \\
    \begin{verbatim}
    {output_format}
    \end{verbatim}
    \end{tcolorbox}
    \caption{LLM-as-judge prompt for evaluating candidate responses against a gold standard using a 1--4 quality scale.}
    \label{fig:candidate_response_evaluation}
\end{figure*}

%% file: figures/prompts/feedback.tex
\begin{figure*}[h]
    \centering
    \tcbset{colframe=black, colback=gray!10, arc=5mm, title={Candidate Response Regeneration with Feedback Prompt}}
    \begin{tcolorbox}
    \scriptsize
    You are provided with a detailed \textbf{video description}, an \textbf{instruction}, a gold standard \textbf{response} for a \textbf{video understanding task} (such as \textbf{captioning}, \textbf{question answering}, \textbf{instruction following}, or \textbf{event description}), and a set of generated \textbf{responses}, each intended to match a specific \textbf{quality rating} from \textbf{4} to \textbf{1}. However, some generated \textbf{responses} were evaluated and found to deviate from their intended quality levels. \\
    
    \textbf{Task:} \\
    Your task is to regenerate revised \textbf{responses} only for those entries where the absolute difference between the \textbf{intended rating} and the \textbf{evaluation rating} is greater than zero (i.e., $|\texttt{intended\_rating} - \texttt{eval\_rating}| > 0$). For each such entry, produce a revised \textbf{response} that strictly conforms to the \textbf{intended quality rating}, based on the definitions in the \textbf{rating guidelines} below. Use the \textbf{video description} as the factual grounding for determining what content is valid. Adjust the \textbf{response} so that its \textbf{evaluation rating} would now exactly match the \textbf{intended rating}. \\
    
    - If the \textbf{eval\_rating} is higher than the \textbf{intended\_rating}, degrade the \textbf{response} by introducing errors such as \textbf{hallucinations}, \textbf{factual distortions}, \textbf{vagueness}, or \textbf{grammar issues}. \\
    - If the \textbf{eval\_rating} is lower than the \textbf{intended\_rating}, improve the \textbf{response} by clarifying actions, reducing errors, or restoring key context from the \textbf{video description}. \\
    
    \textbf{Rating Guidelines:} \\
    - \textbf{Rating 4}: Mostly accurate with minor issues. Preserves key meaning with small lapses. \\
    - \textbf{Rating 3}: Partially correct. Conveys the general idea but with noticeable errors/omissions. \\
    - \textbf{Rating 2}: Poor alignment. Serious flaws, loosely related to task or video. \\
    - \textbf{Rating 1}: Unrelated or incorrect. Fails to reflect the video or follow the instruction. \\
    
    \textbf{Example:} \\
    \textbf{Instruction:} ``Can you give a brief summary of the video content?'' \\
    \textbf{Video Description:} ``A large group of people are participating in an aerobics class inside a spacious indoor venue...'' \\
    \textbf{Gold Standard Response (Rating 5):} ``The video is about a group of people doing a step exercise class, with some confusion among the participants. The focus is on the main woman, who stops at the end and walks towards the camera.'' \\
    \textbf{Rating 4:} ``The video shows a group of people in a step exercise class, with some participants appearing confused during the workout. The main woman stops exercising at the end and moves toward the camera.'' \\
    \textbf{Rating 3:} ``The video features people doing an aerobics class with stepping movements and some coordination issues. There's a woman who seems to be leading the group, and she approaches the camera at some point during the session.'' \\
    \textbf{Rating 2:} ``The video shows several people in what appears to be a dance or fitness routine taking place indoors. A woman in the group walks around and comes closer to where the video is being filmed, while others continue moving in the background.'' \\
    \textbf{Rating 1:} ``The video depicts a cooking demonstration where a chef wearing workout clothes explains different healthy recipes to a small audience in what looks like a kitchen or dining area, with people standing around tables.'' \\
    
    \textbf{Output Format:} \\
    Your final output must be a valid \textbf{JSON object}. For each entry where $|\texttt{intended\_rating} - \texttt{eval\_rating}| > 0$, include a key \texttt{"rating\_\{i\}"} and update the value with a newly revised \textbf{response} that adheres precisely to the \textbf{intended quality rating}. If the \textbf{eval\_rating} is equal to the \textbf{intended\_rating}, do not modify or include that entry. Only output the revised \textbf{responses}. \\
    
    \textbf{Input:} \\
    \textbf{Instruction:} \begin{verbatim}{instruction}\end{verbatim}
    \textbf{Video Description:} \begin{verbatim}{video_description}\end{verbatim}
    \textbf{Gold Standard Response (Rating 5):} \begin{verbatim}{gold_standard_response}\end{verbatim}
    \textbf{Feedback Data (JSON):} \begin{verbatim}{feedback_data}\end{verbatim}
    
    \vspace{2mm}
    \textbf{Output (according to the JSON schema):} \\
    \begin{verbatim}
    {output_format}
    \end{verbatim}
    \end{tcolorbox}
    \caption{LLM-as-judge prompt for regenerating responses to align with intended quality ratings.}
    \label{fig:regeneration_with_feedback}
\end{figure*}

%% file: figures/prompts/train_eval_pointwise.tex
\begin{figure*}[h]
    \centering
    \tcbset{colframe=black, colback=gray!10, arc=5mm, title={Evaluating Model-Generated Responses Prompt}}
    \begin{tcolorbox}
    \scriptsize
    You are provided with a \textbf{video}, a corresponding \textbf{instruction}, and a \textbf{response} generated by a model. 
    The instruction defines a \textbf{video understanding task}, which may take any form --- including but not limited to \textbf{open-ended question answering}, \textbf{captioning}, \textbf{instruction following}, \textbf{temporal reasoning}, or \textbf{multi-step inference grounded in the video}. 
    These tasks are open-ended and often require complex or nuanced reasoning over the visual and temporal content. \\

    Your task is to \textbf{evaluate the quality of the response}, considering how well it satisfies the task defined by the instruction, based on the content of the video. 
    This is a holistic judgment and should be based on the overall \textbf{correctness}, \textbf{relevance}, \textbf{completeness}, and \textbf{grounding} of the response. \\

    \textbf{Task:} \\
    For each response: \\
    - Assess how well it addresses the task in the instruction in the context of the video. \\
    - Consider whether the response is accurate, relevant, complete, and grounded in the video. \\
    - Provide a brief rationale explaining the overall quality and alignment of the response with the instruction inside \texttt{<thinking> </thinking>}. \\
    - Output a score from 1 (worst) to 5 (best) indicating the overall quality inside \texttt{<score> </score>}. \\

    \textbf{Rating Guidelines:} \\
    - \textbf{Rating 5}: Fully accurate, complete, and well-grounded. The response precisely follows the instruction with no notable errors or omissions. \\
    - \textbf{Rating 4}: Mostly accurate with minor issues. The response preserves the key meaning and relevance, with only small lapses in detail or precision. \\
    - \textbf{Rating 3}: Partially correct. The response conveys the general idea but includes noticeable errors, omissions, or misinterpretations. \\
    - \textbf{Rating 2}: Poor alignment. The response has serious flaws and only loosely relates to the instruction or video. \\
    - \textbf{Rating 1}: Unrelated or incorrect. The response fails to reflect the video or follow the instruction meaningfully. \\

    \textbf{Input:} \\
    \textbf{Instruction:} \begin{verbatim}{instruction}\end{verbatim}
    \textbf{Response:} \begin{verbatim}{response}\end{verbatim}

    \vspace{2mm}
    \textbf{Output (Strict Format):} \\
    \begin{verbatim}
    <thinking>{{your reasoning and explanation for the rating}}</thinking>
    <score>{{integer score from 1 to 5}}</score>
    \end{verbatim}
    \end{tcolorbox}
    \caption{LLM-as-judge prompt for evaluating model-generated responses to video understanding tasks.}
    \label{fig:train_eval_pointwise}
\end{figure*}

%% file: figures/prompts/rubric_gen.tex
\begin{figure*}[h]
    \centering
    \tcbset{colframe=black, colback=gray!10, arc=5mm, title={Instruction-Specific Rubrics Generation Prompt}}
    \begin{tcolorbox}
    \scriptsize
    You are provided with a \textbf{video (or a detailed description)}, a corresponding \textbf{instruction}, and a \textbf{reference response}. 
    The instruction defines a \textbf{video understanding task}, which may involve \textbf{open-ended question answering}, \textbf{captioning}, 
    \textbf{instruction following}, \textbf{temporal reasoning}, or \textbf{multi-step inference grounded in the video}. 
    Such tasks are open-ended and often require complex or nuanced reasoning over the visual and temporal content. \\

    Your task is to \textbf{generate an instruction-specific evaluation rubric}. \\
    The rubric will be used to rate the quality of any response to the instruction on a \textbf{1–5 scale}. \\
    The \textbf{reference response is provided only to help you define what a perfect answer (Rating 5) looks like}, 
    but during evaluation the rubric must stand alone — evaluators will not be given the reference response. \\

    \textbf{Important Note:} \\
    Read the provided \textbf{video description as if you are watching the video yourself}. \\
    Pay close attention to all details in the video description and the reference response. \\
    Use these to construct precise, example-specific scoring rubrics that can be applied without needing the reference later. \\

    \textbf{Task:} \\
    For the given video, instruction, and reference response: \\
    - Generate a \textbf{single 1–5 scoring rubric} tailored to this instruction. \\
    - Each score level (1 through 5) must include a clear, example-specific description of what a response at that level would look like. \\
    - Use the \textbf{reference response} to anchor what counts as a "5 (Excellent)" answer. \\
    - Ensure that the rubric is \textbf{self-contained} so that it can be applied without access to the reference response. \\

    \textbf{Rating Guidelines:} \\
    - \textbf{Rating 1 (Very Poor):} Completely wrong, irrelevant, or missing. \\
    - \textbf{Rating 2 (Poor):} Vague, incomplete, or largely inaccurate with minimal grounding. \\
    - \textbf{Rating 3 (Fair):} Partially correct, captures some aspects but misses key details. \\
    - \textbf{Rating 4 (Good):} Mostly correct and grounded, covers most important aspects but not fully comprehensive. \\
    - \textbf{Rating 5 (Excellent):} Fully correct, detailed, coherent, and well-grounded — aligned with the reference response. \\

    \textbf{Input:} \\
    \textbf{Video Content (as detailed description):} \begin{verbatim}{video_description}\end{verbatim}
    \textbf{Instruction:} \begin{verbatim}{instruction}\end{verbatim}
    \textbf{Gold Standard Reference Response:} \begin{verbatim}{reference_response}\end{verbatim}

    \vspace{2mm}
    \textbf{Output (Strict Format):} \\
    \begin{verbatim}
    **Rubric (Scale 1–5):**
    - **Rating 1 (Very Poor):** [instruction-specific description]
    - **Rating 2 (Poor):** [instruction-specific description]
    - **Rating 3 (Fair):** [instruction-specific description]
    - **Rating 4 (Good):** [instruction-specific description]
    - **Rating 5 (Excellent):** [instruction-specific description]
    \end{verbatim}
    \end{tcolorbox}
    \caption{LLM-as-judge prompt for generating instruction-specific rubrics for video understanding tasks.}
    \label{fig:rubric_generation}
\end{figure*}

%% file: tables/rubric_examples.tex
\begin{table*}[t]
\centering
\scriptsize
\setlength{\tabcolsep}{4pt}
\renewcommand{\arraystretch}{1.2}
\caption{Rubrics generated by different models for the instruction \emph{``What is the man wearing while climbing the rock?''}.}
\begin{tabular}{lp{0.78\textwidth}}
\toprule
\textbf{Instruction} & What is the man wearing while climbing the rock? \\
\textbf{Reference Response} & The man is wearing a grey west while climbing the rock. \\
\toprule
\textbf{Model} & \textbf{Rubric} \\
\toprule
Qwen2.5-VL-3B & **Rubric (Scale 1–5):**  
- **Rating 1 (Very Poor):** The response does not address the instruction at all or provides irrelevant information.  
- **Rating 2 (Poor):** The response partially addresses the instruction but contains significant inaccuracies or irrelevant details.  
- **Rating 3 (Fair):** The response addresses the instruction but lacks detail or is not entirely accurate.  
- **Rating 4 (Good):** The response addresses the instruction adequately and provides relevant information, but there are some inaccuracies or omissions.  
- **Rating 5 (Excellent):** The response addresses the instruction thoroughly and accurately, providing all necessary details and information. \\
\midrule
Qwen2.5-VL-7B & **Rubric (Scale 1–5):**  
- **Rating 1 (Very Poor):** The response does not mention the man climbing the rock and instead describes a different scene or activity.  
- **Rating 2 (Poor):** The response mentions the man climbing the rock but includes irrelevant details or descriptions that do not match the video content.  
- **Rating 3 (Fair):** The response accurately describes the man climbing the rock but may include minor inaccuracies or omissions.  
- **Rating 4 (Good):** The response accurately describes the man climbing the rock and includes relevant details, but there might be some minor inaccuracies or omissions.  
- **Rating 5 (Excellent):** The response accurately describes the man climbing the rock, includes all relevant details, and does not contain any inaccuracies or omissions. \\
\midrule
Qwen2.5-VL-32B & **Rubric (Scale 1–5):**  
- **Rating 1 (Very Poor):** The response is completely unrelated to the video content and does not mention any clothing or climbing gear. It may describe an entirely different scene or activity.  
- **Rating 2 (Poor):** The response mentions some clothing or gear, but it is incorrect or irrelevant to the video. For example, it might describe a bright red helmet or a beach setting, which are not present in the video.  
- **Rating 3 (Fair):** The response partially describes the clothing or gear worn by the man climbing the rock, but it is incomplete or contains minor inaccuracies. For example, it might mention a shirt and shorts but fail to note the climbing harness or shoes.  
- **Rating 4 (Good):** The response accurately describes most of the clothing and gear worn by the man climbing the rock, including a shirt, shorts, climbing harness, and shoes. However, it may miss minor details or fail to mention the climbing rope.  
- **Rating 5 (Excellent):** The response provides a complete and accurate description of all the clothing and gear worn by the man climbing the rock, including a shirt, shorts, climbing harness, shoes, and the climbing rope. It is precise and fully grounded in the video content. \\
\midrule
Qwen2.5-VL-72B & **Rubric (Scale 1–5):**  
- **Rating 1 (Very Poor):** The response does not mention any clothing or equipment the man is wearing while climbing the rock. It fails to describe the climbing activity and instead provides irrelevant details about a beach scene.  
- **Rating 2 (Poor):** The response mentions some clothing or equipment but includes significant inaccuracies or irrelevant details. It may describe the wrong activity or setting, such as sitting on a rock or being on a beach.  
- **Rating 3 (Fair):** The response partially describes the man's attire while climbing but misses key details. It may mention some correct elements but lacks completeness or accuracy in describing the climbing gear and setting.  
- **Rating 4 (Good):** The response accurately describes the man's attire, including a tank top, shorts, and climbing shoes, while climbing the rock. It may also mention the climbing harness and rope but could be slightly lacking in detail or context.  
- **Rating 5 (Excellent):** The response comprehensively and accurately describes the man's attire, including a tank top, shorts, climbing shoes, a harness, and a rope, while climbing the rock. It provides a clear and detailed account of the climbing activity and setting, matching the video content perfectly. \\
\midrule
GPT-4o-mini & **Rubric (Scale 1–5):**  
- **Rating 1 (Very Poor):** The response completely fails to address the instruction, providing irrelevant information about the climbing scene or omitting any mention of the man's clothing.  
- **Rating 2 (Poor):** The response offers vague or inaccurate information about the man's clothing, such as a general description (e.g., “he is wearing something”) but fails to specify or properly identify the clothing item.  
- **Rating 3 (Fair):** The response provides a partially correct description of the man's clothing but may miss critical details, such as the specific color or type of clothing, leading to an incomplete answer (e.g., “the man is wearing a shirt”).  
- **Rating 4 (Good):** The response accurately identifies the man's clothing with most important details provided, such as the color and type of the clothing, but might lack some additional context or minor details (e.g., “the man is wearing a grey tank top”).  
- **Rating 5 (Excellent):** The response is fully correct, detailed, and coherent, clearly stating that the man is wearing a grey tank top while climbing the rock, and aligns perfectly with the instruction. \\
\midrule
VideoJudge-3B & **Rubric (Scale 1–5):**  
- **Rating 1 (Very Poor):** The response completely misses the question about what the man is wearing while climbing, providing irrelevant information or no answer at all.  
- **Rating 2 (Poor):** The response gives a vague or incomplete answer regarding the man's clothing, such as mentioning he is “dressed” or “wearing something,” but lacks specific details or accuracy.  
- **Rating 3 (Fair):** The response provides some correct information about the man's clothing but misses key details, such as stating he is “wearing shorts” without specifying the color or type of clothing.  
- **Rating 4 (Good):** The response accurately describes the man's clothing while climbing, mentioning specific items like “a tank top and shorts,” but may lack additional context or detail about his overall appearance.  
- **Rating 5 (Excellent):** The response fully and accurately describes the man's clothing while climbing, clearly stating “the man is wearing a tank top and shorts,” and may also include additional relevant details about his appearance or actions. \\
\bottomrule
\end{tabular}

\label{tab:rubrics_example}
\end{table*}

%% file: figures/prompts/train_eval_pointwisewithrubric.tex
\begin{figure*}[h]
    \centering
    \tcbset{colframe=black, colback=gray!10, arc=5mm, title={Rubric Generation and Response Evaluation Prompt}}
    \begin{tcolorbox}
    \scriptsize
    You are provided with a \textbf{video}, a corresponding \textbf{instruction}, and a \textbf{response} generated by a model. 
    The instruction defines a \textbf{video understanding task}, which may take any form --- including but not limited to 
    \textbf{open-ended question answering}, \textbf{captioning}, \textbf{instruction following}, \textbf{temporal reasoning}, 
    or \textbf{multi-step inference grounded in the video}. 
    These tasks are open-ended and often require complex or nuanced reasoning over the visual and temporal content. \\

    Your task has two parts: \\

    \textbf{Part 1 — Rubric Generation:} \\
    - Generate an \textbf{instruction-specific evaluation rubric} on a \textbf{1–5 scale} for this example. \\
    - The rubric must define what a Rating 1 through Rating 5 response would look like \textbf{in the context of this exact video and instruction}. \\
    - The rubric should be \textbf{tailored to this instruction} and \textbf{grounded in the video content}, not generic. \\
    - Each rating should describe concrete aspects of the video/instruction that would or would not appear in a response at that level. \\
    - The rubric should be \textbf{self-contained} so it can later be applied to any response to the same instruction, without needing additional references. \\
    - Output your rubric within \texttt{<rubric></rubric>}. \\

    \textbf{Part 2 — Response Evaluation:} \\
    - Using the rubric you just generated, evaluate the provided response. \\
    - Assess how well it addresses the task in the instruction in the context of the video. \\
    - Consider whether the response is accurate, relevant, complete, and grounded in the video. \\
    - Provide a brief rationale explaining the overall quality and alignment of the response with the instruction inside \texttt{<thinking> </thinking>}. \\
    - Output a score from 1 (worst) to 5 (best) indicating the overall quality inside \texttt{<score> </score>}. \\

    \textbf{Input:} \\
    \textbf{Instruction:} \begin{verbatim}{instruction}\end{verbatim}
    \textbf{Response:} \begin{verbatim}{response}\end{verbatim}

    \vspace{2mm}
    \textbf{Output (Strict Format):} \\
    \begin{verbatim}
    <rubric>
    **Rubric (Scale 1–5):**
    - **Rating 1 (Very Poor):** [instruction-specific description referencing video details]
    - **Rating 2 (Poor):** [instruction-specific description referencing video details]
    - **Rating 3 (Fair):** [instruction-specific description referencing video details]
    - **Rating 4 (Good):** [instruction-specific description referencing video details]
    - **Rating 5 (Excellent):** [instruction-specific description referencing video details]
    </rubric>
    <thinking>{{your reasoning and explanation for the rating}}</thinking>
    <score>{{integer score from 1 to 5}}</score>
    \end{verbatim}
    \end{tcolorbox}
    \caption{LLM-as-judge prompt for generating rubrics and evaluating responses to video understanding tasks.}
    \label{fig:train_eval_pointwisewithrubric}
\end{figure*}

%% file: figures/prompts/rubric_eval.tex
\begin{figure*}[h]
    \centering
    \tcbset{colframe=black, colback=gray!10, arc=5mm, title={Rubric Comparison Prompt}}
    \begin{tcolorbox}
    \scriptsize
    You are provided with a \textbf{video}, a corresponding \textbf{instruction}, and a \textbf{reference response}. 
    The instruction defines a \textbf{video understanding task}, which may take any form --- including but not limited to 
    \textbf{open-ended question answering}, \textbf{captioning}, \textbf{instruction following}, \textbf{temporal reasoning}, 
    or \textbf{multi-step inference grounded in the video}. 
    These tasks are open-ended and often require complex or nuanced reasoning over the visual and temporal content. \\

    You are also provided with \textbf{two rubrics (Rubric A and Rubric B)}, each generated by a model. 
    These rubrics are intended to evaluate responses to the instruction on a \textbf{1–5 scale}. \\

    Your task is to \textbf{decide which rubric is better} for evaluating responses to the instruction. 
    This is a holistic judgment and it should be based on the rubric’s \textbf{specificity}, \textbf{clarity}, \textbf{coverage}, and \textbf{usefulness for evaluation}. \\

    \textbf{Task:} \\
    For each pair of rubrics: \\
    - Assess which rubric better reflects the instruction and video content (instruction-specificity). \\
    - Consider whether the rubric is self-contained and usable without the reference response. \\
    - Evaluate the clarity, distinctness, and logical progression of the rating levels (1–5). \\
    - Judge which rubric provides better coverage of key aspects required to evaluate responses. \\
    - Provide a brief rationale explaining your choice inside \texttt{<thinking> </thinking>}. \\
    - Output the preferred rubric as A or B inside \texttt{<answer> </answer>}. \\

    \textbf{Guidelines for Comparison:} \\
    - Prefer the rubric that is \textbf{more specific} to the given instruction and video. \\
    - Prefer the rubric that is \textbf{clear, well-structured, and easy to apply}. \\
    - Prefer the rubric that \textbf{captures all important aspects} of what makes a good or bad response. \\
    - If both rubrics are strong, choose the one that is \textbf{slightly more precise or comprehensive}. \\
    - Do not output a tie — always select either A or B. \\

    \textbf{Input:} \\
    \textbf{Instruction:} \begin{verbatim}{instruction}\end{verbatim}
    \textbf{Reference Response:} \begin{verbatim}{ref_response}\end{verbatim}

    \textbf{Rubric A:} \begin{verbatim}{rubric_a}\end{verbatim}
    \textbf{Rubric B:} \begin{verbatim}{rubric_b}\end{verbatim}

    \vspace{2mm}
    \textbf{Output (Strict Format):} \\
    \begin{verbatim}
    <thinking>{{your reasoning and explanation for why one rubric is better}}</thinking>
    <answer>{{A or B}}</answer>
    \end{verbatim}
    \end{tcolorbox}
    \caption{LLM-as-judge prompt for pairwise comparison of two instruction-specific rubrics.}
    \label{fig:rubric_comparison}
\end{figure*}

%% file: figures/prompts/train_eval_pairwise.tex
\begin{figure*}[h]
    \centering
    \tcbset{colframe=black, colback=gray!10, arc=5mm, title={Pairwise Response Comparison Prompt}}
    \begin{tcolorbox}
    \scriptsize
    You are provided with a \textbf{video}, a corresponding \textbf{instruction}, and \textbf{two candidate responses} generated by two models. 
    The instruction defines a \textbf{video understanding task}, which may involve \textbf{open-ended question answering}, \textbf{captioning}, 
    \textbf{instruction following}, \textbf{temporal reasoning}, or \textbf{multi-step inference grounded in the video}. 
    Such tasks are open-ended and often require complex or nuanced reasoning over the visual and temporal content. \\

    Your task is to \textbf{compare the two responses} and decide which one better addresses the instruction, 
    based on the content of the video. 
    Make a holistic judgment that considers \textbf{correctness}, \textbf{relevance}, \textbf{completeness}, and \textbf{grounding}. \\

    \textbf{Task:} \\
    For the given pair of responses: \\
    - Judge which response better addresses the instruction in the context of the video. \\
    - Consider whether each response is accurate, relevant, complete, and grounded in the video. \\
    - Output only \texttt{A} or \texttt{B}, wrapped strictly inside \texttt{<answer></answer>} tags. \\

    \textbf{Evaluation Guidelines:} \\
    - \textbf{Accuracy}: Prefer responses that are factually correct and consistent with the video. \\
    - \textbf{Relevance}: Prefer responses that directly answer the instruction without digression. \\
    - \textbf{Completeness}: Prefer responses that capture all key aspects needed for a full answer. \\
    - \textbf{Grounding}: Prefer responses clearly supported by the video, avoiding hallucinations. \\
    - If one response contains hallucinations, irrelevant content, or omissions, prefer the other. \\
    - If both responses are strong, choose the one that is more precise and detailed. \\
    - If both responses are weak, choose the one that is less flawed. \\

    \textbf{Input:} \\
    \textbf{Instruction:} \begin{verbatim}{instruction}\end{verbatim}

    \textbf{Response A:} \begin{verbatim}{response_a}\end{verbatim}
    \textbf{Response B:} \begin{verbatim}{response_b}\end{verbatim}

    \vspace{2mm}
    \textbf{Output (Strict Format):} \\
    \begin{verbatim}
    <answer>{{A_or_B}}</answer>
    \end{verbatim}
    \end{tcolorbox}
    \caption{LLM-as-judge prompt for pairwise comparison of model-generated responses to video understanding tasks.}
    \label{fig:train_eval_pairwise}
\end{figure*}

%% file: figures/prompts/train_eval_pairwisewithfeedback.tex
\begin{figure*}[h]
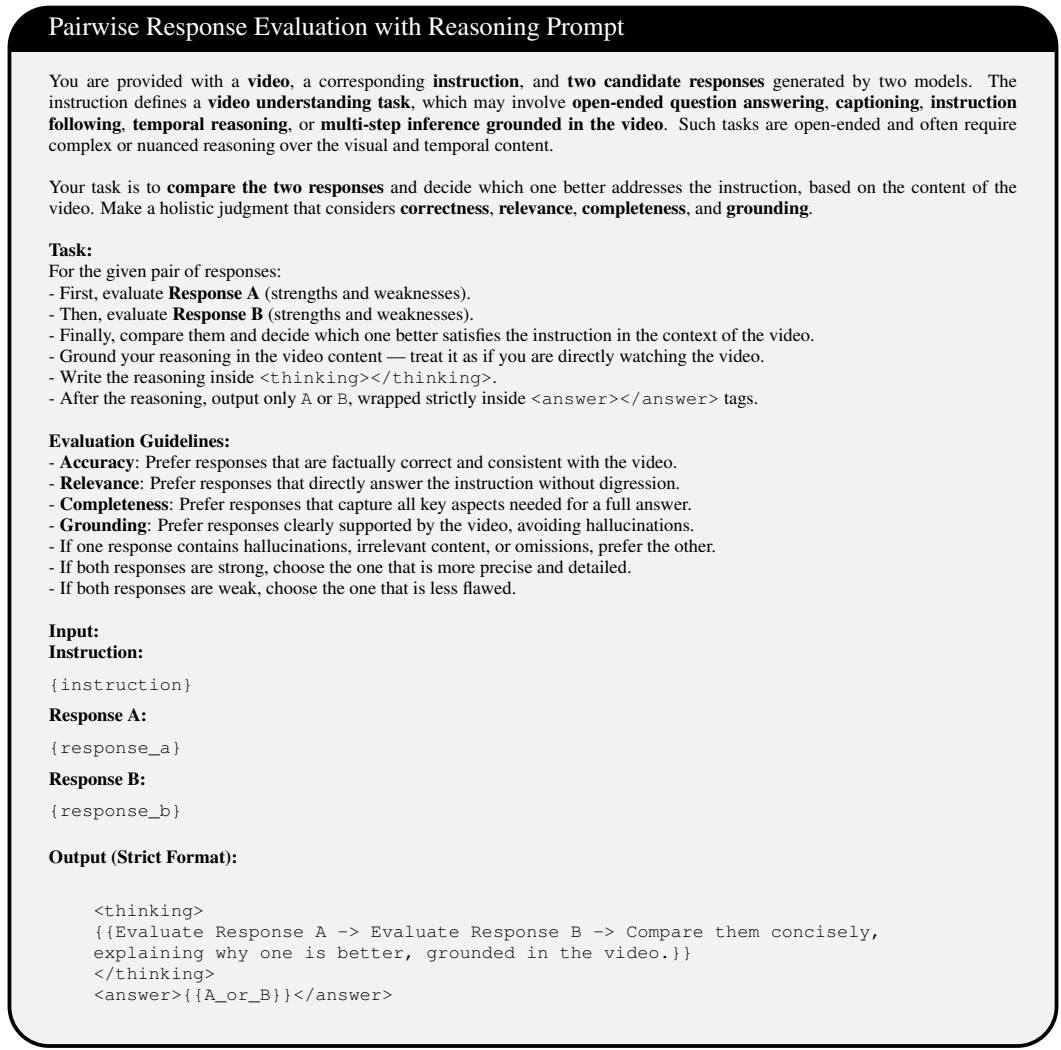

    \centering
    \tcbset{colframe=black, colback=gray!10, arc=5mm, title={Pairwise Response Evaluation with Reasoning Prompt}}
    \begin{tcolorbox}
    \scriptsize
    You are provided with a \textbf{video}, a corresponding \textbf{instruction}, and \textbf{two candidate responses} generated by two models. 
    The instruction defines a \textbf{video understanding task}, which may involve \textbf{open-ended question answering}, \textbf{captioning}, 
    \textbf{instruction following}, \textbf{temporal reasoning}, or \textbf{multi-step inference grounded in the video}. 
    Such tasks are open-ended and often require complex or nuanced reasoning over the visual and temporal content. \\

    Your task is to \textbf{compare the two responses} and decide which one better addresses the instruction, 
    based on the content of the video. Make a holistic judgment that considers \textbf{correctness}, \textbf{relevance}, 
    \textbf{completeness}, and \textbf{grounding}. \\

    \textbf{Task:} \\
    For the given pair of responses: \\
    - First, evaluate \textbf{Response A} (strengths and weaknesses). \\
    - Then, evaluate \textbf{Response B} (strengths and weaknesses). \\
    - Finally, compare them and decide which one better satisfies the instruction in the context of the video. \\
    - Ground your reasoning in the video content --- treat it as if you are directly watching the video. \\
    - Write the reasoning inside \texttt{<thinking></thinking>}. \\
    - After the reasoning, output only \texttt{A} or \texttt{B}, wrapped strictly inside \texttt{<answer></answer>} tags. \\

    \textbf{Evaluation Guidelines:} \\
    - \textbf{Accuracy}: Prefer responses that are factually correct and consistent with the video. \\
    - \textbf{Relevance}: Prefer responses that directly answer the instruction without digression. \\
    - \textbf{Completeness}: Prefer responses that capture all key aspects needed for a full answer. \\
    - \textbf{Grounding}: Prefer responses clearly supported by the video, avoiding hallucinations. \\
    - If one response contains hallucinations, irrelevant content, or omissions, prefer the other. \\
    - If both responses are strong, choose the one that is more precise and detailed. \\
    - If both responses are weak, choose the one that is less flawed. \\

    \textbf{Input:} \\
    \textbf{Instruction:} \begin{verbatim}{instruction}\end{verbatim}

    \textbf{Response A:} \begin{verbatim}{response_a}\end{verbatim}
    \textbf{Response B:} \begin{verbatim}{response_b}\end{verbatim}

    \vspace{2mm}
    \textbf{Output (Strict Format):} \\
    \begin{verbatim}
    <thinking>
    {{Evaluate Response A -> Evaluate Response B -> Compare them concisely, 
    explaining why one is better, grounded in the video.}}
    </thinking>
    <answer>{{A_or_B}}</answer>
    \end{verbatim}
    \end{tcolorbox}
    \caption{LLM-as-judge prompt for pairwise evaluation of responses with explicit stepwise reasoning.}
    \label{fig:train_eval_pairwisewithfeedback}
\end{figure*}

%% file: sections/algorithm.tex
\begin{algorithm}[h]
\caption{Bootstrapping Training Data with Self-Refinement}
\label{alg:bootstrap}
\KwIn{Video $v$, instruction $x$, gold response $y^{*}$, generator $G$, evaluator $E$, threshold $\alpha$, max iterations $T$}
\KwOut{Bootstrapped dataset $\mathcal{D}$}

Initialize $\mathcal{D} \leftarrow \{(v, x, y^{*}, N)\}$ \tcp{gold response with max rating}

\For{$r \in \{1, \dots, N-1\}$}{
    $y^{(r)}_{0} \leftarrow G(p_{\text{gen}} \Vert v \Vert x \Vert y^{*}, r)$ \tcp{initial generation}
    
    \For{$t \in \{0, \dots, T-1\}$}{
        $\hat{r}, f^{(r)}_{t} \leftarrow E(p_{\text{eval}} \Vert v \Vert x \Vert y^{*} \Vert y^{(r)}_{t})$ \\
        \If{$|r - \hat{r}| \leq \alpha$}{
            $\mathcal{D} \leftarrow \mathcal{D} \cup \{(v, x, y^{(r)}_{t}, r)\}$ \\
            \textbf{break}
        }
        \Else{
            $y^{(r)}_{t+1} \leftarrow G(p_{\text{ref}} \Vert v \Vert x \Vert y^{*} \Vert y^{(r)}_{t} \Vert f^{(r)}_{t}, r)$
        }
    }
}
\Return{$\mathcal{D}$}
\end{algorithm}

%% file: tables/bootstrapped_data_example.tex
\begin{table*}[ht!]
\centering
\scriptsize
\setlength{\tabcolsep}{4pt}
\renewcommand{\arraystretch}{1.15}
\caption{Representative examples of video frames paired with instructions and bootstrapped responses at different rating levels (R1–R5) generated by our pipeline.}
\begin{tabularx}{\textwidth}{@{}>{\centering\arraybackslash}m{0.36\textwidth} >{\raggedright\arraybackslash}X@{}}
\toprule
\textbf{Video Frames} & \textbf{Instruction and Responses} \\
\midrule

% --- Example 1 ---
\begin{minipage}[t]{\linewidth}
  \centering
  \vspace{0.1cm}
  \includegraphics[width=\linewidth]{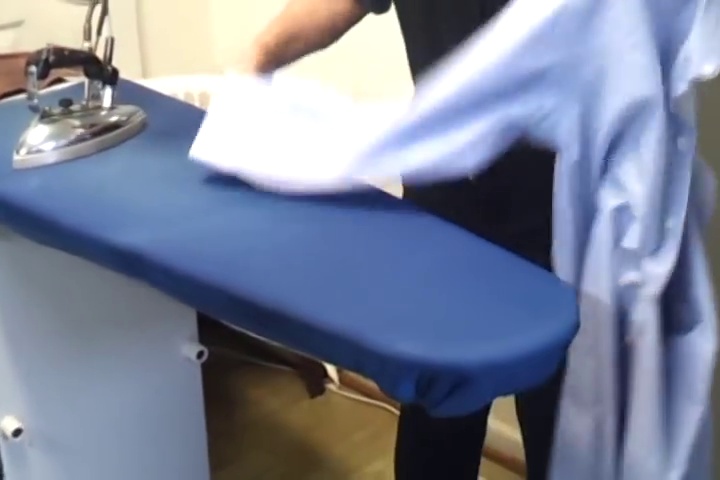}\\[0.6ex]
  \includegraphics[width=\linewidth]{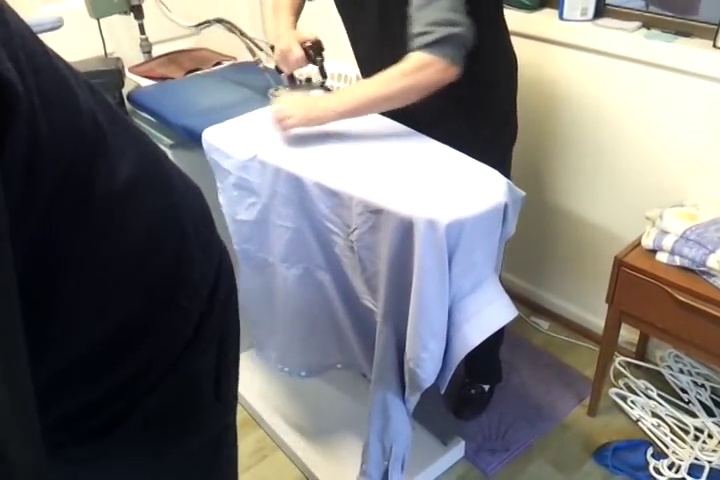}
\end{minipage}
&
\begin{minipage}[t]{\linewidth}
\textbf{Instruction:} What is the man wearing while ironing the dress shirt?

\medskip
\textbf{R5:} The video shows a white building from the outside, the inside of a ballet studio where children practice, a girl speaking from inside a well-lit studio, children walking through the school, and a boy and a girl dancing in a studio as other children watch.  

\smallskip
\textbf{R4:} The video features an exterior view of a grand building, followed by scenes inside a ballet studio where young dancers practice. There are moments showing a girl speaking and other children walking through the school, along with a boy and a girl performing ballet moves while others observe.  

\smallskip
\textbf{R3:} The video features a beautiful building at the start, then transitions into a ballet class where children are practicing their routines. A girl talks at one point, and there are some glimpses of kids walking around. The main focus is on the ballet practice, with some children watching and learning.  

\smallskip
\textbf{R2:} The video shows a building and then cuts to some kids who are doing ballet in a studio. There is a girl who talks briefly, and some kids are just wandering around. The dancing is mixed with scenes of children doing other activities, which makes it hard to tell what's happening.  

\smallskip
\textbf{R1:} The video features a beautiful building at the start, then transitions into a ballet class where children are practicing their routines. A girl talks at one point, and there are some glimpses of kids walking around. The main focus is on the ballet practice, with some children watching and learning.
\end{minipage}
\\
\midrule

% --- Example 2 ---
\begin{minipage}[t]{\linewidth}
  \centering
  \vspace{0.1cm}
  \includegraphics[width=\linewidth]{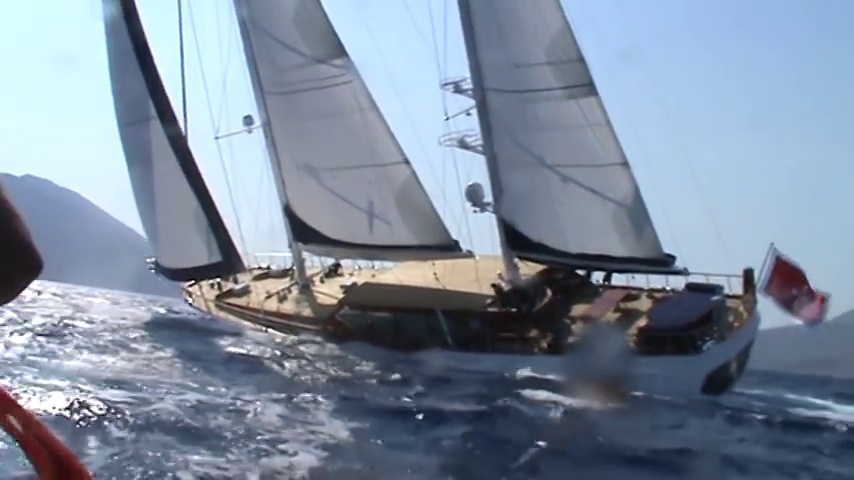}\\[0.6ex]
  \includegraphics[width=\linewidth]{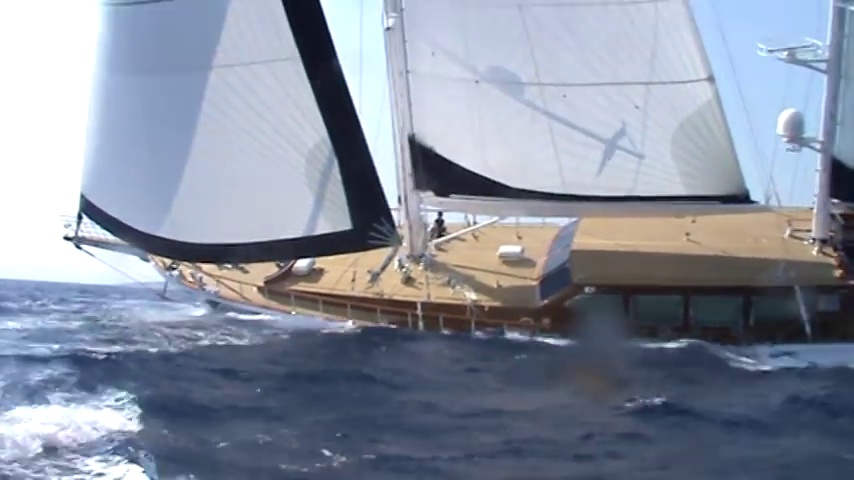}
\end{minipage}
&
\begin{minipage}[t]{\linewidth}
\textbf{Instruction:} Can you describe in detail what happens in the video?

\medskip
\textbf{R5:} The video depicts a ship in the sea during the daytime, with a hill visible in the distance at the shore. The ship is dangerously tilted towards the left and appears to be on the verge of capsizing, while all its sails are fully deployed. Another group of people can be seen approaching the ship in a motorboat. The video ends with the camera panning around the motorboat and the surrounding area. There are no visible signs of distress or emergency response. Overall, the video captures a dramatic and potentially hazardous situation on the high seas.  

\smallskip
\textbf{R4:} The video shows a sailing yacht navigating through rough waters during the day, with a distant coastline visible. The yacht is leaning significantly to one side, giving the impression that it might capsize, and all its sails are up. A motorboat approaches the yacht, but there are no signs of distress. The scene captures an adventurous sailing experience in open waters.  

\smallskip
\textbf{R3:} The video features a yacht sailing in the ocean on a sunny day, with some hills visible far in the background. The yacht is leaning to one side, which could look dangerous. All the sails are fully deployed, and it appears to be a thrilling experience. There seems to be no immediate danger, and the overall atmosphere feels calm despite the tilt of the yacht, making it look like a leisurely outing rather than an emergency.  

\smallskip
\textbf{R2:} The video presents a boat moving across the water, and it's bright outside. There are some distant landforms, possibly hills. The boat seems to be leaning, which might suggest it's having trouble. A small boat is around, but it’s unclear what the situation is. The video feels like it’s capturing a sailing adventure.  

\smallskip
\textbf{R1:} The video shows a large ship floating in calm waters, with people enjoying a picnic on the deck. In the background, there's a beautiful sunset, and the ship appears to be stationary. The focus is on people laughing and eating, with no signs of sailing or any movement.
\end{minipage}
\\
\bottomrule
\end{tabularx}

\label{tab:bootstrapped_examples}
\end{table*}

%% file: tables/hyperparameters.tex
\begin{table}[ht!]
\centering
\footnotesize
\setlength{\tabcolsep}{8pt}
\renewcommand{\arraystretch}{1.15}
\caption{Training and evaluation hyperparameters.}

\begin{tabular}{@{}ll@{}}
\toprule
\textbf{Hyperparameter} & \textbf{Value} \\
\midrule
\multicolumn{2}{@{}l}{\textit{Training}} \\[0.5ex]
Learning rate & 2e-7 \\
Batch size (per device) & 16 \\
Gradient accumulation steps & 1 \\
Num.\ train epochs & 2 \\
Warmup ratio & 0.03 \\
Weight decay & 0.0 \\
Max grad norm & 1.0 \\
LR scheduler & Cosine \\
Precision & bfloat16 \\
Max sequence length & 128{,}000 \\
Video max frames & 60 \\
Video max frame pixels & 25{,}088 \\
Video min frame pixels & 3{,}136 \\
Gradient checkpointing & True \\
tune\_mm\_vision & False \\
tune\_mm\_mlp & True \\
tune\_mm\_llm & True \\
\midrule
\multicolumn{2}{@{}l}{\textit{Evaluation}} \\[0.5ex]
max\_new\_tokens & 1024 \\
fps & 1 \\
max\_frames & 180 \\
max\_pixels & 20480 $\times$ 28 $\times$ 28 \\
min\_pixels & 16 $\times$ 28 $\times$ 28 \\
\bottomrule
\end{tabular}
\label{tab:hyperparams}
\end{table}

%% file: tables/pairwise_humaneval.tex
\begin{table}[ht]
\centering
\setlength{\tabcolsep}{6pt}
\renewcommand{\arraystretch}{1.2}
\caption{Pairwise human evaluation results across two annotators. 
The table reports overall inter-annotator agreement and Cohen's Kappa as measures of reliability. 
Preference distributions show the proportion of times each annotator selected response \texttt{b} versus response \texttt{a}, indicating a slight bias toward \texttt{b}. 
Correctness is computed as the fraction of instances where the annotator’s preferred response matches the higher-rated (gold) response, with both annotators showing high accuracy over 250 samples each. 
The last two rows capture error analysis: in 11 cases (4.4\%), both annotators agreed on an incorrect answer, while in 13 cases (5.2\%), the annotators disagreed on a correct answer, highlighting residual uncertainty.}
\begin{tabular}{lcc}
\toprule
\textbf{Metric} & \textbf{Value} \\
\midrule
Agreement & 94.80 \\
Cohen's Kappa & 89.54 \\
\midrule
Annotator1 Preference (b / a) & 53.2 / 46.8 \\
Annotator2 Preference (b / a)   & 54.4 / 45.6 \\
\midrule
Annotator1 Correctness & 92.40 (250 samples) \\
Annotator2 Correctness   & 93.60 (250 samples) \\
\midrule
Both agreed on wrong answers & 11 / 250 (4.4) \\
Both disagreed on correct answers & 13 / 250 (5.2) \\
\bottomrule
\end{tabular}

\label{tab:humaneval_results}
\end{table}

%% file: tables/pairwise_humaneval_examples.tex
\begin{table*}[t]
\centering
\scriptsize
\setlength{\tabcolsep}{3pt}
\renewcommand{\arraystretch}{1.1}
\caption{Comparison of annotator decisions across instruction--response pairs.}

\begin{tabular}{p{2.5cm} p{4.0cm} p{4.0cm} p{0.8cm} p{0.8cm}}
\toprule
\textbf{Instruction} & \textbf{Response A} & \textbf{Response B} & \textbf{A1} & \textbf{A2} \\
\midrule
What is happening in the video? & The video shows a couple dancing outside, with some people watching them. Children are playing in the background, and there might be someone cooking food. Another couple appears later, but the details are a bit unclear. & In the video, a couple is moving around in what looks like a backyard gathering. There are kids playing, and it seems like a party. A few adults are standing around, but it's hard to see what's really happening. & A & A \\
Can you describe the video in detail? & In the video, a man in a camouflage shirt is seen doing nail care for a woman. He appears to be using some kind of product, and there are people sitting nearby. The lighting seems good, and the focus is on the nails, but there are moments where it shifts away from the main action. & The video features a casual setting where a person is doing something with nails. There is a man in a camouflage shirt, and it looks like he is applying some kind of treatment. Other people are around, but the details about the process are quite vague and unclear. & B & A \\
Can you describe the competition the man is participating in? & The video shows a man participating in a sports competition where he throws a round object in a circular area. He attempts to throw it a long distance, and there are some spectators watching, but the focus on the throwing technique is not very clear, and the details about the event are vague. & In the video, a man is seen throwing a discus in a competition. He spins to gain speed and then throws it. The footage captures him from different perspectives, and there are some people in the crowd, but not many details about them. & A & B \\
What is the woman wearing and what is behind her? & The woman has on a green shirt, and there seems to be a clothing item behind her, but it's not specific. The room is bright, and there might be some products nearby, but the details are unclear. & The woman is wearing a green top, and behind her is a denim jacket. She is in a well-lit room with a window, and there are some products on a table, but it's not clear what specific items they are. & B & B \\
\bottomrule
\end{tabular}
\label{tab:annotator_comparison}
\end{table*}

%% file: figures/videojudge_winrate_additional.tex
\begin{figure*}[t]
    \centering
    \begin{subfigure}[t]{0.48\textwidth}
        \centering
        \includegraphics[width=\linewidth]{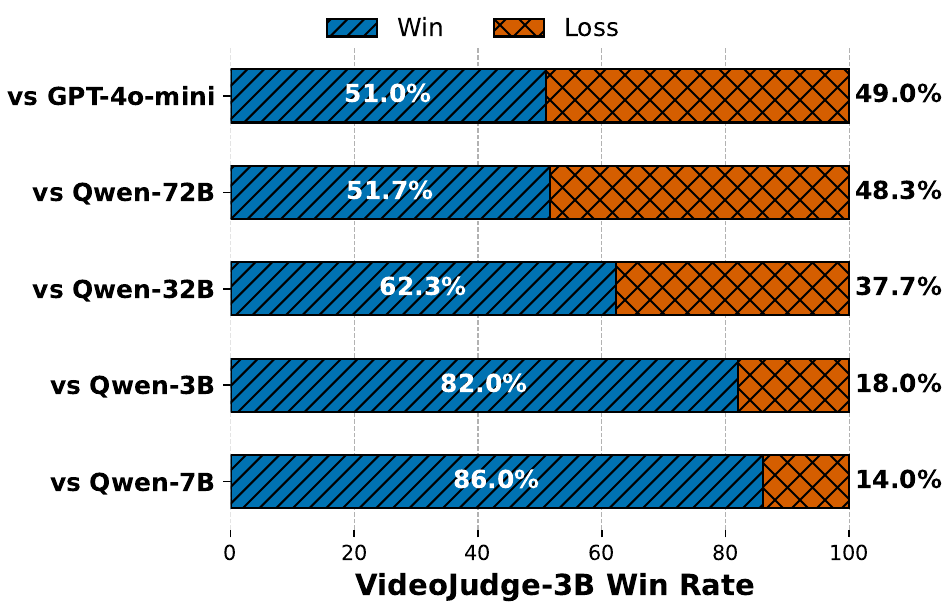}
        \caption{Majority vote with 3 annotators}
        \label{fig:videojudge_winrate_majority}
    \end{subfigure}
    \hspace{0.01\textwidth} % small controlled gap
    \begin{subfigure}[t]{0.48\textwidth}
        \centering
        \includegraphics[width=\linewidth]{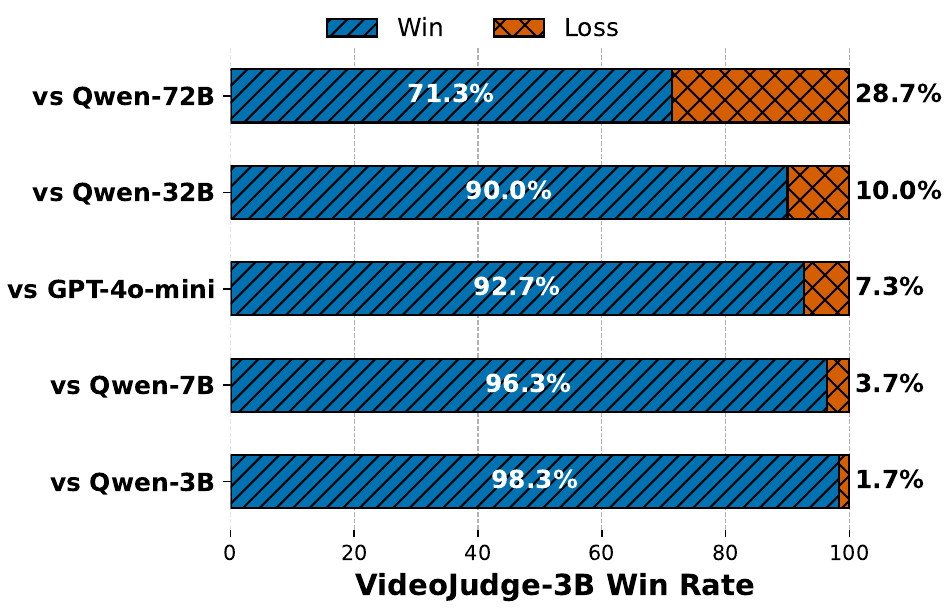}
        \caption{LLM-as-Judge based evaluation}
        \label{fig:llm_as_judge_preference}
    \end{subfigure}
    \caption{ Win rate of \textbf{VideoJudge-3B} compared to other models under two evaluation settings. \ref{fig:videojudge_winrate_majority}: majority voting of human annotator, where the most common annotators' choice determines the label. \ref{fig:llm_as_judge_preference}: LLM-as-Judge preference with deterministic decoding ($T=0$). Across both settings, \textbf{VideoJudge-3B} consistently produces high-quality rubrics and achieves performance competitive with, or surpassing, models up to 25$\times$ larger, including proprietary systems such as GPT-4o-mini. }
    \label{fig:videojudge_winrate_additional}
\end{figure*}

%% file: figures/temperature_others.tex
\begin{figure*}[t]
    \centering
    \begin{subfigure}[t]{0.32\textwidth}
        \centering
        \includegraphics[width=\linewidth]{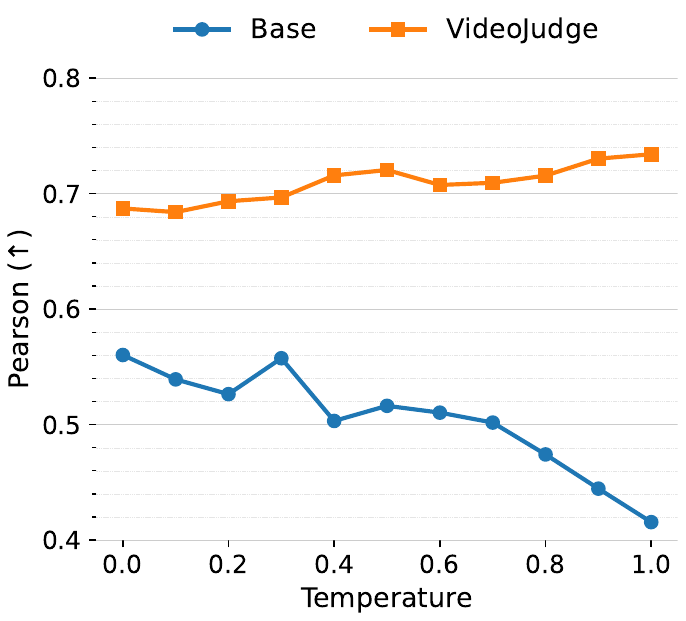}
        \caption{Pearson}
        \label{fig:pearson}
    \end{subfigure}
    \hfill
    \begin{subfigure}[t]{0.32\textwidth}
        \centering
        \includegraphics[width=\linewidth]{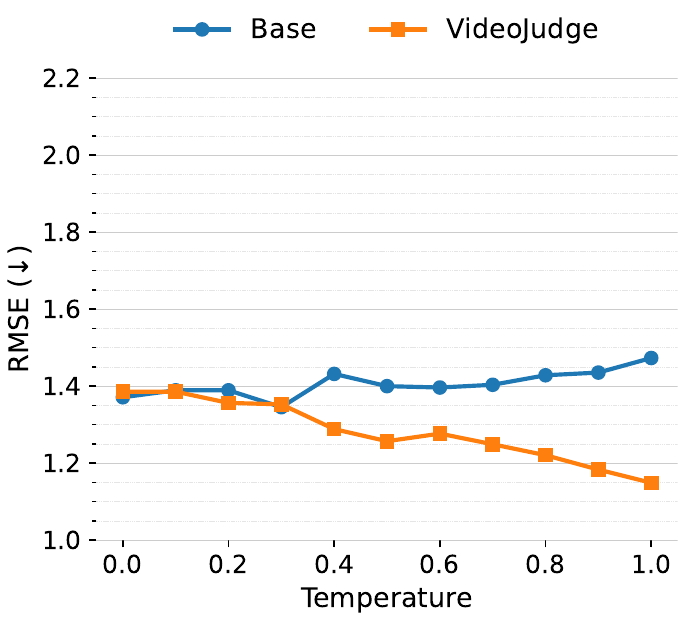}
        \caption{RMSE}
        \label{fig:rmse}
    \end{subfigure}
    \hfill
    \begin{subfigure}[t]{0.32\textwidth}
        \centering
        \includegraphics[width=\linewidth]{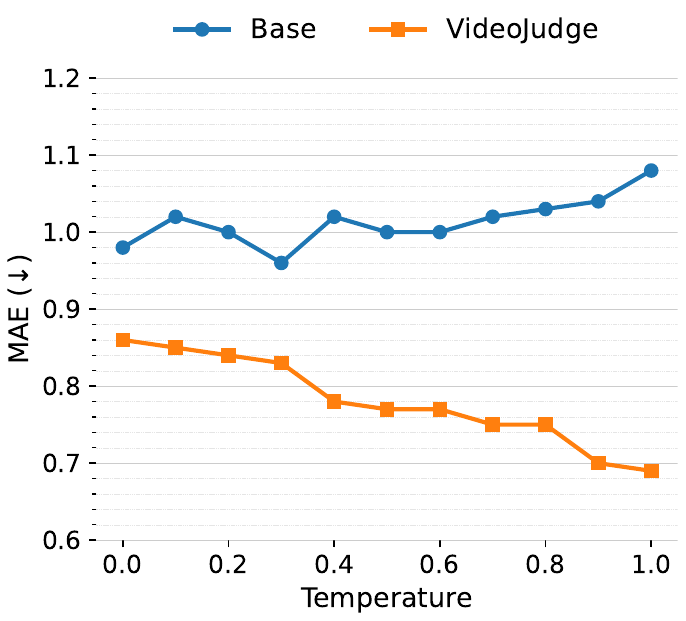}
        \caption{MAE}
        \label{fig:mae}
    \end{subfigure}
    \caption{Comparison of Zero-Shot vs Finetuned models across temperatures using Pearson, RMSE, and MAE metrics.}
    \label{fig:temperature_others}
\end{figure*}

%% file: figures/maxframes_others.tex
\begin{figure*}[t]
    \centering
    % ---------------- Training ----------------
    \begin{subfigure}{0.32\linewidth}
        \centering
        \includegraphics[width=\linewidth]{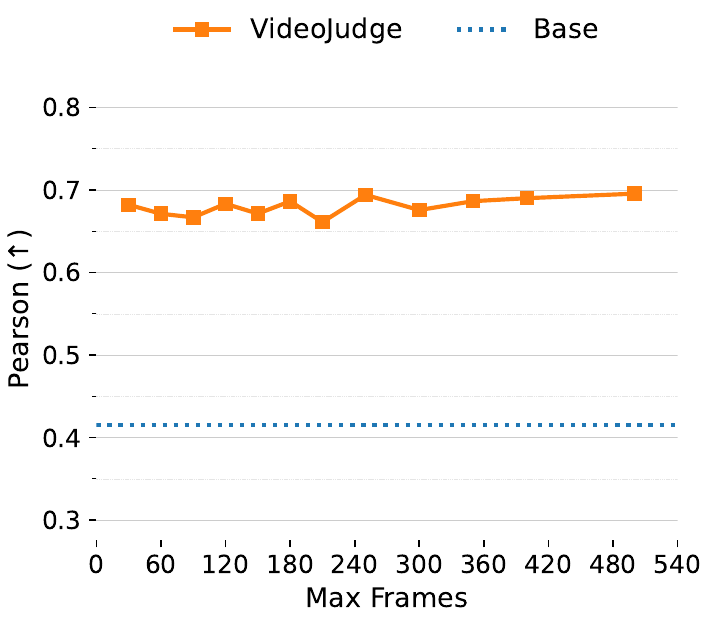}
        \caption{Pearson (Train).}
        \label{fig:maxframes_train_pearson}
    \end{subfigure}
    \hfill
    \begin{subfigure}{0.32\linewidth}
        \centering
        \includegraphics[width=\linewidth]{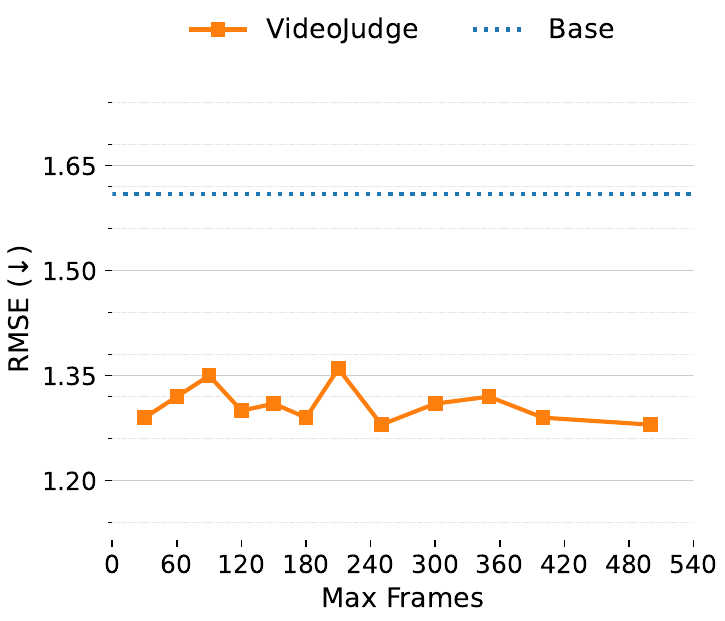}
        \caption{RMSE (Train).}
        \label{fig:maxframes_train_rmse}
    \end{subfigure}
    \hfill
    \begin{subfigure}{0.32\linewidth}
        \centering
        \includegraphics[width=\linewidth]{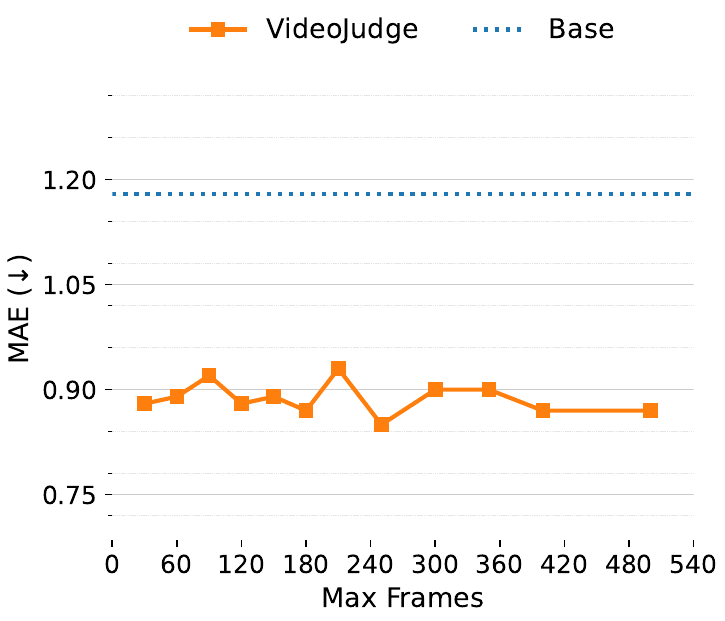}
        \caption{MAE (Train).}
        \label{fig:maxframes_train_mae}
    \end{subfigure}

    \vspace{1.2ex}

    % ---------------- Evaluation ----------------
    \begin{subfigure}{0.32\linewidth}
        \centering
        \includegraphics[width=\linewidth]{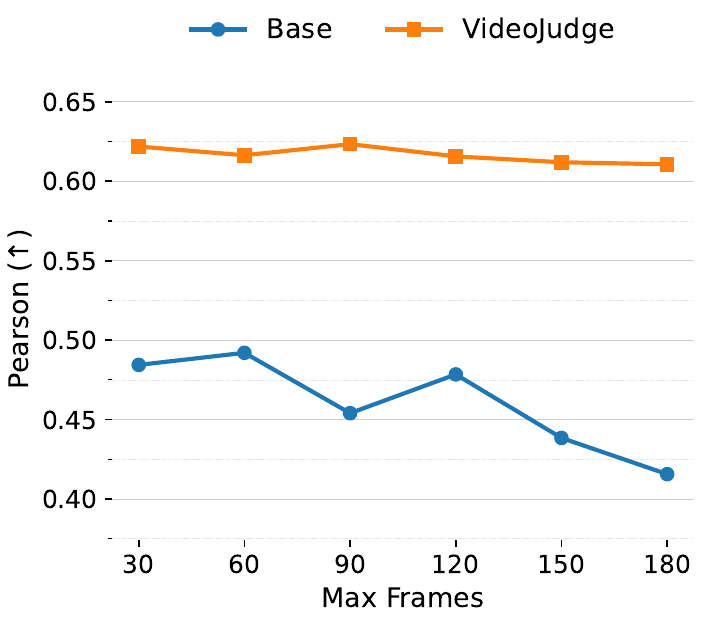}
        \caption{Pearson (Eval).}
        \label{fig:maxframes_eval_pearson}
    \end{subfigure}
    \hfill
    \begin{subfigure}{0.32\linewidth}
        \centering
        \includegraphics[width=\linewidth]{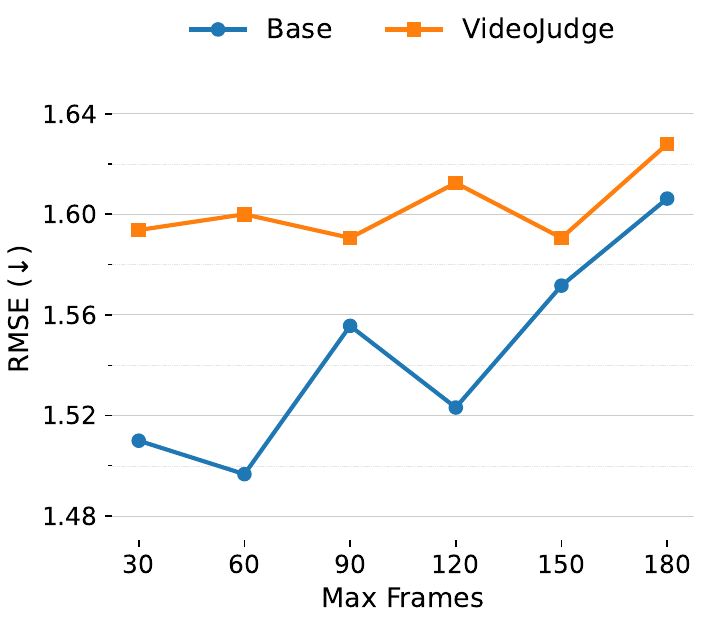}
        \caption{RMSE (Eval).}
        \label{fig:maxframes_eval_rmse}
    \end{subfigure}
    \hfill
    \begin{subfigure}{0.32\linewidth}
        \centering
        \includegraphics[width=\linewidth]{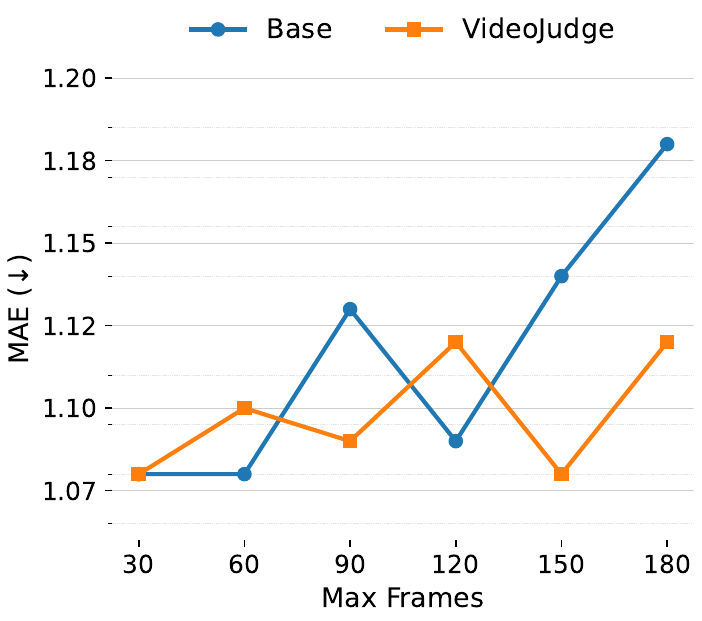}
        \caption{MAE (Eval).}
        \label{fig:maxframes_eval_mae}
    \end{subfigure}

    \vspace{-1ex}
    \caption{Training vs. evaluation results across Pearson, RMSE, and MAE metrics for max-frame ablation.}
    \label{fig:maxframes_others}
\end{figure*}